\documentclass[twoside,11pt]{article}

\usepackage{blindtext}

\usepackage{jmlr2e}
\usepackage{booktabs}
\usepackage{tablefootnote}
\usepackage{dsfont}
\usepackage{amsmath}
\usepackage{multirow}
\usepackage{enumitem}
\usepackage{colortbl}
\usepackage{fancybox}
\usepackage{listings}
\usepackage{xcolor}

\usepackage{bm}
\usepackage{pifont}   
\newcommand{\xmark}{\textcolor{red}{\ding{55}}}      
\newcommand{\cmark}{\textcolor{green}{\ding{51}}}    
\usepackage{subcaption} 
\usepackage{tabularx}
\usepackage{wrapfig}

\usepackage[utf8]{inputenc} 
\usepackage[T1]{fontenc}    

\usepackage{lastpage}
\jmlrheading{26}{2025}{1-\pageref{LastPage}}{12/24; Revised
10/25}{10/25}{24-2141}{Jianguo Huang, Jianqing Song, Xuanning Zhou, Bingyi Jing, Hongxin Wei}
\ShortHeadings{TorchCP: A Python Library for Conformal Prediction }{Huang, Song, Zhou, Jing and Wei}

\firstpageno{1}

\begin{document}

\title{TorchCP: A Python Library for Conformal Prediction}

\author{\centering
  Jianguo Huang$^{1,2}$, 
  Jianqing Song$^{1,3}$, 
  Xuanning Zhou$^{1,4}$,\\
  Bingyi Jing$^{1,4}$, 
  Hongxin Wei$^{1}$\thanks{Corresponding author: Hongxin Wei (weihx@sustech.edu.cn)}\\
\normalfont{
\begin{tabular}{l}
$^1$Southern University of Science and Technology \quad
$^2$Nanyang Technological University \\
$^3$Nanjing University \quad
$^4$The Chinese University of Hong Kong, Shenzhen
\end{tabular}
}
}



\editor{Zeyi Wen}

\def\torchcp{{\fontfamily{cmss}{\fontseries{l}\selectfont TorchCP}}\xspace}

\maketitle

\begin{abstract}

Conformal prediction (CP) is a powerful statistical framework that generates prediction intervals or sets with guaranteed coverage probability. While CP algorithms have evolved beyond traditional classifiers and regressors to sophisticated deep learning models like deep neural networks (DNNs), graph neural networks (GNNs), and large language models (LLMs), existing CP libraries often lack the model support and scalability for large-scale deep learning (DL) scenarios. This paper introduces \torchcp, a PyTorch-native library designed to integrate state-of-the-art CP algorithms into DL techniques, including DNN-based classifiers/regressors, GNNs, and LLMs.
Released under the LGPL-3.0 license, \torchcp comprises about 16k lines of code, validated with 100\% unit test coverage and detailed documentation. Notably, \torchcp enables CP-specific training algorithms, online prediction, and GPU-accelerated batch processing, achieving up to 90\% reduction in inference time on large datasets. With its low-coupling design, comprehensive suite of advanced methods, and full GPU scalability, \torchcp empowers researchers and practitioners to enhance uncertainty quantification across cutting-edge applications.

\end{abstract}


\section{Introduction}

Conformal prediction (CP) is a statistical framework generating prediction intervals or sets with guaranteed coverage probability~\citep{f60917dcc9144362b788a01b03e6894e}, offering a distribution-free solution to quantify predictive uncertainty. 
While CP algorithms have been developed beyond traditional classifiers and regressors to sophisticated deep learning (DL) models like graph neural networks (GNNs) and large language models (LLMs), existing CP libraries often lack the model support and scalability required for large-scale DL scenarios. This underscores the need for a comprehensive and scalable CP toolbox tailored for DL models, facilitating robust and accessible conformal prediction methods across cutting-edge applications.

This work presents \torchcp, a PyTorch-native library that advances conformal prediction for DL by implementing a wide range of state-of-the-art CP algorithms, including training and inference methods tailored for models such as DNN classifiers/regressors, GNNs, and LLMs. Its low-coupling design facilitates easy customization of components like trainers or scores, while full GPU support and batch processing ensure seamless scalability for large-scale tasks. \torchcp excels over existing libraries by delivering superior speed, with up to 90\% faster inference times on large datasets compared to \texttt{MAPIE} and \texttt{PUNCC} (Appendix~\ref{appendix:comparsion_libraries}), while also encompassing a broader algorithmic coverage than \texttt{TorchUQ}. This PyTorch-native integration and extensibility empower statisticians and DL practitioners to seamlessly apply CP across cutting-edge applications with enhanced DL compatibility.

\vspace{-8pt}
\section{Main contributions}

Conformal prediction constructs a prediction set $C(X)$ for an input $X$, ensuring the true outcome $Y$ satisfies $P(Y\in\mathcal{C}(X))\geq1-\alpha$, where $ \alpha $ is a miscoverage rate. Existing CP libraries encounter significant challenges in DL, including limited native support for modern models (like DNNs, GNNs, and LLMs), poor computational efficiency when scaling to large datasets, and monolithic designs that restrict the flexible customization for diverse applications. \torchcp overcomes these shortcomings through a PyTorch-native framework that improves model compatibility, efficiency, and adaptability, thereby enhancing CP's applicability to DL. Its low-coupling design and GPU-accelerated batch processing provide a robust foundation for tackling complex, real-world problems, detailed in the following parts.


\begin{wrapfigure}{r}{0.4\textwidth}
    \vspace{-1.5em}  
    \centering
    \includegraphics[width=0.4\textwidth]{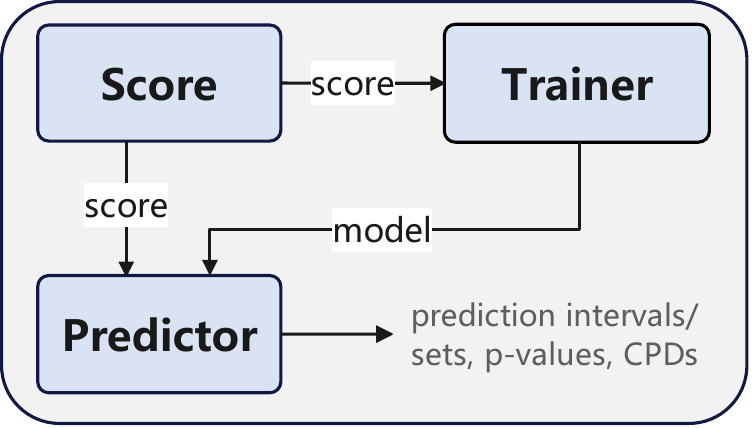}
    \caption{The modules of \torchcp.}
    \label{fig:modular_design}
    \vspace{-13pt}  
\end{wrapfigure}


\subsection{Advantages for DL scenarios}


\textbf{Low-coupling design} Low-coupling design is one of the most fundamental concepts in object-oriented design~\citep{mcnatt2001coupling}, which minimizes the dependency a class has on other classes. While many libraries (e.g., \texttt{PUNCC}) are implemented in a high-coupling design, TorchCP is structured into three core modules: Trainer, Score, and Predictor, each tailored to specific aspects of CP workflows. The \textbf{Trainer} module implements CP-specific training algorithms, such as ConfTr \citep{stutz2021learning}, which optimizes deep models to improve efficiency. The \textbf{Score} module calculates non-conformity measures, e.g., LAC~\citep{sadinle2019least}, to quantify how atypical a new sample is relative to the training data. The \textbf{Predictor} module implements CP workflows, including calibration and prediction, to generate outputs, e.g., prediction sets/intervals, p-values, and conformal predictive distributions.



This low-coupling design is visualized in Figure~\ref{fig:modular_design}, illustrating the interplay among modules. 
The Score module is used to create various score objects that serve as inputs for both the Trainer and Predictor modules.
The Trainer module optimizes DL models using CP-specific loss functions. 
The Predictor module then integrates the trained model with the score objects to generate conformal outputs, such as prediction sets/intervals, p-values, or predictive distributions. In this manner, \torchcp achieves strong reusability and extensibility by providing standardized module interfaces, making it straightforward to define new methods. Appendix~\ref{appendix:extensibility} provides guidelines for adding customized modules using the provided interfaces, while Appendix~\ref{appendix:dis_puncc} discusses the advantages of \torchcp’s low-coupling design.

\begin{table}[!t]
\centering
\small
\begin{tabular}{p{2.7cm} p{2.3cm} p{9.5cm}}  
\toprule
\textbf{Model types} & \textbf{Component} & \textbf{CP Algorithms} \\
\midrule
\multirow{3}{*}{\raisebox{-8\height}{ \textbf{DNN Classifier}}} 
& Trainer\&Loss & ConfTr~\citep{stutz2021learning}, ConfTS~\citep{xi2024does}, C-Adapter~\citep{liu2024c}, UA Loss~\citep{einbinder2022training}, SCPO~\citep{bellotti2021optimized} \\
\cmidrule{2-3}
& Score Function & 
LAC~\citep{sadinle2019least}, APS~\citep{romano2020classification}, RAPS~\citep{angelopoulos2020uncertainty},
SAPS~\citep{huangconformal},
KNN~\citep{gammerman2007hedging}, Margin~\citep{lofstrom2015bias},  TOPK~\citep{angelopoulos2020uncertainty} \\
\cmidrule{2-3}
& Predictor & 
Split CP~\citep{papadopoulos2002inductive},Class-conditional CP~\citep{vovk2012conditional}, Clustered CP~\citep{ding2023class}, Weighted CP~\citep{tibshirani2019conformal}, RC3P~\citep{shi2024conformal} \\
\midrule
\multirow{3}{*}{\raisebox{-5\height}{ \textbf{DNN Regressor}} }
& Score Function & ABS~\citep{papadopoulos2002inductive}, 
NORABS~\citep{papadopoulos2011regression}, 
CQR~\citep{romano2019conformalized}, 
CQRR~\citep{sesia2020comparison}, 
CQRM~\citep{sesia2020comparison}, 
CQRFM~\citep{kivaranovic2020adaptive}, 
R2CCP~\citep{guha2023conformal}\\
 \cmidrule{2-3}
& Predictor & Split CP,  Ensemble CP~\citep{xu2021conformal}, 
ACI~\citep{gibbs2021adaptive}, 
AgACI~\citep{zaffran2022adaptive}\\
\midrule
\multirow{3}{*}{\raisebox{-1.5\height}{\textbf{GNN}}} & Trainer\&Loss  & 
CF-GNN~\citep{huang2024uncertainty} \\
 \cmidrule{2-3}
& Score Function &  DAPS~\citep{zargarbashi2023conformal}, 
SNAPS~\citep{song2024similarity}  \\
 \cmidrule{2-3}
& Predictor &  NAPS~\citep{clarkson2023distribution} \\
\midrule
\textbf{LLM} & Predictor & Conformal LLM~\citep{quachconformal} \\
\bottomrule
\end{tabular}
\caption{The CP algorithms implemented in \torchcp for various deep learning models.}
\label{tab:torchcp_scope}
\vspace{-11pt}
\end{table}

\vspace{-5pt}
\paragraph{Methodological scope}
Table~\ref{tab:torchcp_scope} outlines the CP algorithms in \torchcp across 4 types of DL models\footnote{Model types are defined by their input/output data differences, impacting the choice of CP algorithms.}: DNN classifiers/regressors, GNNs, and LLMs, along with evaluation metrics in Table~\ref{tab:torchcp_evaluation_metrics}. Current libraries are typically limited in algorithmic coverage, e.g., \texttt{MAPIE} only includes split CP. In \torchcp, we provide 6 training algorithms, 17 scores, and 10 predictors.
A brief overview of all supported methods and their advantages is available in Appendix~\ref{appendix:all_methods}. 

\paragraph{Batch Processing and GPU Support} \torchcp is a PyTorch-native library equipped with GPU acceleration and batch processing capabilities. It enables efficient computation on large datasets and scales seamlessly to DL applications, e.g., achieving up to 90\% reduction in inference time on ImageNet (See Figure~\ref{fig:imagenet_consuming_time_diff}). Appendix~\ref{appendix:efficiency} presents empirical evidence of its efficiency advantages on large-scale datasets. 

\vspace{-8pt}
\subsection{Tutorials}
\torchcp provides a user-friendly framework for conformal prediction in deep learning with a detailed documentation\footnote{https://torchcp.readthedocs.io/en/latest/index.html}. The library contains illustrative code showcasing functionality for common applications (See the ``examples'' folder~\footnote{\url{https://github.com/ml-stat-Sustech/TorchCP/tree/master/examples}}) with a practical guide in Appendix~\ref{appendix:exmaple_code}.
Moreover, in Appendix~\ref{appendix:results}, we provide reproduced results of classical CP methods, confirming the correctness of our implementation. Below, we present a generic example~\footnote{The full source code is available at \href{https://github.com/ml-stat-Sustech/TorchCP/blob/master/examples/classification_conftr_cifar100.py}{TorchCP CIFAR-100 Example}.} to show the usability of \torchcp. First, we utilize ConfTr~\citep{stutz2021learning} to train a neural network, i.e., \texttt{init\_model}, on a training dataset encapsulated in \texttt{train\_loader}:\\

\begin{lstlisting}
from torchcp.classification.trainer import ConfTrTrainer
trainer = ConfTrTrainer(init_model, alpha=0.1) # Initialize trainer with model  
trained_model = trainer.train(train_loader, num_epochs=10) # Training 10 epochs
\end{lstlisting}


With the trained model, users can proceed with conformal prediction by selecting an appropriate score function and predictor.
Below, we implement Split CP with a miscoverage rate of $\alpha=0.1$,  using the LAC score on calibration data encapsulated in \texttt{cal\_loader}:

\begin{lstlisting}
predictor = SplitPredictor(score_function=LAC(), model=trained_model, alpha=0.1)
predictor.calibrate(cal_loader) # Set threshold with calibration data
predict_set = predictor.predict(test_instance) # Predict set for test instance
\end{lstlisting}
\vspace{-10pt}

\section{Related library}
Conformal prediction is a distribution-free framework for quantifying predictive uncertainty, yet existing libraries face notable limitations in DL applications. Libraries, such as \texttt{nonconformist}, \texttt{PUNCC}, and \texttt{MAPIE}, offer restricted native support for modern models like GNNs and LLMs, relying on NumPy without GPU acceleration, which hinders efficient scaling to large datasets. Moreover, tools like \texttt{TorchUQ} and \texttt{Fortuna} provide some GPU support but fall short in algorithmic coverage, e.g., \texttt{TorchUQ} supports only regressors. Additionally, many libraries like \texttt{PUNCC} are implemented in a high-coupling design, where predictors and score functions are tightly coupled within a single class. \torchcp addresses these gaps with its PyTorch-native integration, supporting a broader range of modern models and offering a modular, extensible architecture, as detailed in Table~\ref{tab:summarization}. We present a detailed comparison of the relevant libraries in Appendix~\ref{appendix:comparsion_libraries}.

\begin{table}[!t]
\resizebox{\textwidth}{!}{
\setlength{\tabcolsep}{2mm}{ 
\begin{tabular}{lccccccccccccc}
\hline
    \multirow{2}{*}{\centering Library} & \multicolumn{4}{c}{Model Support}  & \multicolumn{6}{c}{Functionality Support}  & \multirow{2}{*}{\centering Coupling} & \multirow{2}{*}{\centering Backend} \\
    \cmidrule(r){2-5} \cmidrule(r){6-11}  & Classifier & Regressor &  GNN & LLM & Training & Online  & GPU    & Batch & $p$-value & CPD  \\
\midrule
\texttt{nonconformist}~\citep{nonconformist} & \cmark & \cmark  & \xmark & \xmark & \xmark  & \xmark & \xmark & \xmark & \xmark & \xmark & \textcolor{green}{low} & Numpy \\
\texttt{TorchUQ}~\citep{torchuq2024} & \xmark & \cmark  & \xmark & \xmark & \xmark & \cmark & \cmark & \cmark & \xmark & \xmark & \textcolor{green}{low} & PyTorch\\
\texttt{Crepes}~\citep{crepes} & \cmark & \cmark & \xmark & \xmark & \xmark  &  \cmark & \xmark & \xmark & \cmark & \cmark & \textcolor{green}{low} & Numpy\\
\texttt{PUNCC}~\citep{mendil2023puncc} & \cmark & \cmark  & \xmark & \xmark & \xmark & \xmark & \xmark & \xmark & \xmark & \xmark & \textcolor{red}{high} & Numpy\\
\texttt{MAPIE}~\citep{Cordier_Flexible_and_Systematic_2023} & \cmark & \cmark  & \xmark & \xmark & \xmark & \xmark & \xmark & \xmark & \xmark & \cmark & \textcolor{red}{high} & Numpy\\
\texttt{Fortuna}~\citep{detommaso2024fortuna} & \cmark & \cmark  & \xmark & \xmark & \xmark & \xmark & \cmark & \cmark  & \xmark & \xmark & \textcolor{red}{high} & JAX\\
\torchcp (ours) & \cmark & \cmark & \cmark & \cmark & \cmark & \cmark & \cmark & \cmark & \cmark & \cmark & \textcolor{green}{low} & PyTorch \\
\bottomrule
\end{tabular}}}
\caption{Comparison of Open-Source CP Libraries~\tablefootnote{The definitions of model support and functionality support are presented in Appendix~\ref{appendix:comparsion_libraries}.}. ``Classifier'' and ``Regressor'' refer to DNN-based classifiers and regressors\tablefootnote{\torchcp also supports traditional machine learning models like random forests, but is primarily designed for advanced DL models within the PyTorch ecosystem.}. The term ``Training'' denotes support for CP-specific training, while ``Online'' refers to capabilities for semi-online conformal prediction. ``$p$-value'' and ``CPD'' denote outputting $p$-values and conformal predictive distributions, respectively.}\label{tab:summarization}
\vspace{-10pt}
\end{table}

\section{Conclusion and Future Work}
In this work, we introduced \torchcp, a PyTorch-native library that seamlessly integrates conformal prediction algorithms into deep learning workflows. Supporting a wide range of state-of-the-art CP algorithms, \torchcp leverages PyTorch’s GPU acceleration and batch processing, offering a low-coupling design for easy customization and user-friendly documentation for accessibility. Looking ahead, we plan to expand \torchcp by incorporating emerging CP algorithms for evolving architectures like diffusion models and multimodal models, while optimizing for real-time performance. 
Through open-source collaboration on GitHub, we aim to foster community-driven innovation, empowering reliable AI in fields such as healthcare, finance, and autonomous systems.

\acks{Hongxin Wei is supported by the Shenzhen Fundamental Research Program (Grant No. JCYJ20230807091809020). Bingyi Jing's research is partly supported by NSFC 12371290. We gratefully acknowledge
the support of the Center for Computational Science and
Engineering at the Southern University of Science and Technology for our research.}


\bibliography{sample}

\newpage
\appendix

\section{Details of Conformal Prediction Methods in \torchcp}
\label{appendix:all_methods}


In this section, we provide brief introductions to the methods implemented in \torchcp. Furthermore, Table~\ref{tab:cp_methods_adavantages} offers practical guidelines to help users select appropriate methods based on their specific application needs.

\begin{table}[htbp]
\centering
\small
\begin{tabularx}{\textwidth}{@{} l l X @{}}
\toprule
\textbf{Component} & \textbf{Method} & \textbf{Advantage/Use case} \\
\midrule
\rowcolor{gray!20}
\multicolumn{3}{c}{\textbf{DNN Classifier}} \\
\midrule
\multirow{7}{*}{Score} 
  & LAC      & To achieve the smallest prediction sets \\
  & APS      & To approximate conditional coverage \\
  & RAPS     & For smaller, more stable sets than APS \\
  & TOPK     & A simple baseline with only logit ranking\\
  & SAPS     & Improving efficiency, conveying instance-wise uncertainty \\
  & Margin   & A non-conformity function based on prediction margin \\
  & KNN      & A score function based solely on feature-space nearest neighbors without a predictive model. \\
\midrule
\multirow{3}{*}{Predictor}
  & Class-conditional CP & To achieve coverage for each class \\
  & Clustered CP         & For class-conditional coverage with limited data per class \\
  & Weighted CP          & Conformal prediction under covariate shift \\
\midrule
\multirow{3}{*}{Loss \& Trainer}
  & ConfTr     & Efficient prediction sets with conformal-aware loss \\
  & C-Adapter  & Optimizing efficiency based on $\alpha$-free loss \\
  & ConfTS     & Improving efficiency via temperature scaling \\
\midrule
\rowcolor{gray!20}
\multicolumn{3}{c}{\textbf{DNN Regressor}} \\
\midrule
\multirow{3}{*}{Score}
  & ABS      & Simple and interpretable residual-based scores \\
  & NORABS   & Enhancing Prediction Interval Robustness through Normalized Scoring \\
  & CQR      & Flexible interval width via quantile regression \\
    & R2CCP         & Shorter prediction intervals than CQR via recasting regression as classification \\
\midrule
\multirow{4}{*}{Predictor}
  & ACI           & Reliable coverage under distributional shifts \\
  & AgACI         & Aggregates ACI predictions with adaptive weighting \\
  & Ensemble CP   & Stable prediction intervals without data splitting via model ensembling \\

\midrule
\rowcolor{gray!20}
\multicolumn{3}{c}{\textbf{GNN}} \\
\midrule
\multirow{3}{*}{Score}
      & DAPS     & Improves efficiency of APS using structural neighbors \\
  & SNAPS    & Extends DAPS with feature neighbors for better efficiency \\
  & NAPS     & Constructing prediction sets for inductive node classification via Weighted CP \\
\midrule
Loss \& Trainer 
  & CF-GNN   & Reducing set size by attaching an adapter with a topology-aware inefficiency loss\\
\midrule
\rowcolor{gray!20}
\multicolumn{3}{c}{\textbf{LLM}} \\
\midrule
Predictor & Conformal LLM &  To achieve statistically guaranteed prediction sets for language models through calibrated sampling \\
\bottomrule
\end{tabularx}
\caption{Advantages or use cases of CP methods across different tasks.}
\label{tab:cp_methods_adavantages}
\end{table}

\subsection{Split CP}
\label{appendix:split_cp}

Conformal prediction~\citep{papadopoulos2002inductive, balasubramanian2014conformal, manokhin_2022_6467205,angelopoulos2023conformal,angelopoulos2024theoretical} is a statistical framework that generates prediction sets/intervals containing ground-truth labels with a desired probability. \textit{Split Conformal Prediction}~\citep{f60917dcc9144362b788a01b03e6894e} (Split CP) is the most widely-used version of the conformal prediction procedure. Specifically, Split CP divides a given dataset into two disjoint subsets: one for training the base model and the other for conformal calibration.
We next outline the main process of split conformal prediction:

\begin{enumerate}
\item Divide a given dataset into two disjoint subsets: a training fold $\mathcal{D}_{tr}$ and a calibration fold $\mathcal{D}_{cal}$, with $|\mathcal{D}_{cal}|$ being $n$;
\item Train a model on the training dataset $\mathcal{D}_{tr}$, and then define a non-conformity score function $V(\bm{x}, y)$;
\item Compute $\widehat{Q}_{1-\alpha}$ as the $\frac{\lceil(n+1)(1-\alpha)\rceil}{n}$ quantile of the calibration scores $\left \{V(\bm{x}_i,y_i): (\boldsymbol{x}_i,y_i) \in \mathcal{D}_{cal}\right \} $, where $\widehat{Q}_{1-\alpha}$ is defined by
$$ \widehat{Q}_{1-\alpha} := \inf \{ Q\in\mathbb{R}: \frac{ |\{i:V(\bm{x}_i,y_i) \leq Q \}|} {n} \geq  \frac{\lceil  (n+1)(1-\alpha) \rceil}{n}\}; $$
\item Use the threshold $\widehat{Q}_{1-\alpha}$ to generate a prediction set for a new instance $\bm{x}_{n+1}$:
$$\mathcal{C}(\bm{x}_{n+1}) = \{y: V(\bm{x}_{n+1},y)\leq \widehat{Q}_{1-\alpha}\}.$$
\end{enumerate}

\subsection{DNN Classifier}

Here, we provide a brief overview of conformal prediction methods implemented using a DNN classifier with the \torchcp library.
Let $\mathcal{X}\subset\mathbb{R}^{d}$ be the input space and $\mathcal{Y}:=\{1,\dots,K\}$ be the label space. We use $(X,Y)\sim \mathcal{P}_{\mathcal{X}\mathcal{Y}}$ to denote a random data pair satisfying a joint data distribution $\mathcal{P}_{\mathcal{X}\mathcal{Y}}$ and  $f:\mathcal{X} \rightarrow \mathbb{R}^K$ to denote a classification neural network. Thus, the classifier $\pi:\mathcal{X} \rightarrow \Delta^{K-1}$is defined as $\sigma\circ f$, where $\Delta^{K-1}$ is a ($K$-1)-dimensional probability simplex and  $\sigma$ is a normalization function such as the softmax function. Let $\pi_{y}(x)$ be the probability of $y$ conditional on $x$, i.e., $P(y|x)$.


\textbf{LAC \citep{sadinle2019least}}:  The LAC measures the similarity between the example and the data space by the conditional probability $P(Y|X)$: $V(\bm{x},y) := 1- \pi_y(\bm{x})$.

\textbf{APS \citep{romano2020classification}}:  The score of Adaptive Prediction Sets (APS) calculates the non-conformity score using the cumulative sum of sorted softmax probabilities:
$$
 V(\boldsymbol{x},y,u) := \sum\limits_{y^\prime \in\mathcal{Y}} \pi_{y^\prime} (\bm{x})\mathds{1}(r_f(\bm{x},y^\prime)< r_f(\bm{x},y))  + u \cdot \pi_y(\bm{x}),$$
where $r_f(\bm{x},y)$ denotes the rank of $\pi_{y}(\boldsymbol{x})$ among the descending softmax probabilities, and $u$ is an independent random variable satisfying a uniform distribution on $[0,1]$.

\textbf{RAPS \citep{angelopoulos2020uncertainty}}: RAPS adds a regularization to penalize noisy tail probabilities and regularize the number of samples in the uncertainty set:
$$    V(\bm{x},y,u;\pi) :=  \sum\limits_{y^\prime \in\mathcal{Y}} \pi_{y^\prime} (\bm{x})\mathds{1}(r_f(\bm{x},y^\prime)< r_f(\bm{x},y))  + u \cdot \pi_y(\bm{x}) + \lambda \cdot (r_f(\bm{x},y)-k)^{+},$$ 
where $\lambda$ represents the weight of regularization, $k \geq 0$ are regularization hyper-parameters and $(z)^{+}$ denotes the positive part of $z$.

\textbf{TOPK \citep{angelopoulos2020uncertainty}}: TOPK is a score function based solely on the ranking information: $V(\boldsymbol{x},y,u) := \sum\limits_{y^\prime \in\mathcal{Y}} \mathds{1}(r_f(\bm{x},y^\prime)< r_f(\bm{x},y))  + u$.

\textbf{SAPS \citep{huangconformal}}:  The SAPS method mitigates the issue of probability miscalibration by only retaining the highest probability value and discarding all others:
$$ V(\bm{x},y,u) := \left\{ 
      \begin{array}{ll}
        u \cdot\pi_{max} (\bm{x}), \qquad \text{if\quad} r_f(\bm{x},y)=1,\\
        \pi_{max} (\bm{x}) +(r_f(\bm{x},y)-2+u) \cdot \lambda, \quad \text{else},
      \end{array}
    \right.$$
where $\lambda$ is a hyperparameter representing the weight of ranking information and  $\pi_{max} (\boldsymbol{x})$ denotes the maximum softmax probability.

\textbf{KNN \citep{gammerman2007hedging}}: KNN calculates the score based on the k closest training examples in the feature space. The score is calculated by dividing two sums: the numerator is the sum of the k shortest distances from an input point to other points that share its same label, while the denominator is the sum of the k shortest distances from that input point to points that have different labels than it does.

\textbf{Margin \citep{lofstrom2015bias}}: The margin score is constructed on a well-known concept that a prediction with a larger margin is generally considered to be more conforming. Consequently, the margin score is defined by $V{(\bm{x},y)}:= \pi_{y}-\max _{j\in\mathcal{Y}: {j} \neq y} \pi_{j}.$

\textbf{Class-conditional CP \citep{vovk2012conditional}}: Class-conditional CP partitions the calibration data based on their labels $y$. The non-conformity scores' quantiles are then computed separately for each class using the corresponding calibration subset.

\textbf{Clustered CP \citep{ding2023class}}: Clustered CP extends Class-conditional CP by applying split conformal prediction to class clusters rather than individual classes. It achieves cluster-conditional coverage as an approximation of class-conditional guarantees.

\textbf{Weighted CP \citep{tibshirani2019conformal}}: Weighted CP addresses distribution shift in Conformal Prediction, where calibration and test data follow different distributions. The core idea is that data points closer to the test point in the feature space provide more valuable information about it.

\textbf{RC3P~\citep{shi2024conformal}}: RC3P addresses the inefficiency of class-conditional conformal prediction, which often results in large prediction sets, especially for imbalanced classification tasks. The core idea is to selectively perform class-wise thresholding only for labels with low top-k error, by integrating a label rank calibration step into standard class-wise conformal calibration.

\textbf{ConfTr \citep{stutz2021learning}}: ConfTr simulates the CP process based on end-to-end training by splitting mini-batches into calibration and prediction sets and designing differentiable conformal prediction steps:
$\mathcal{L}_{\mathrm{ConfTr}}=\max \left(0, \sum_{k=1}^{K} \mathcal{C}_{\theta, k}(\bm{x})-\kappa\right),$
where $\theta$ is the trainable parameters of deep learning models and $\kappa$ is the target set size.
 The definition of $\mathcal{C}_{\theta, y}(\bm{x}) $ is given by $\sigma ((\widehat{Q}_{1-\alpha}- V(\bm{x},y))/\tau)$ where $\tau$ is a temperature hyperparameter and $\sigma$ is the sigmoid function.

\textbf{ConfTS \citep{xi2024does}}: ConfTS is a training method to optimize the $T$ of temperature scaling for improved efficiency of prediction sets. The objective of ConfTS is to minimize the efficiency gap, i.e., $\widehat{Q}_{1-\alpha}-V(\bm{x},y)$. Formally, the loss function of ConfTS is:  $$\mathcal{L}_{\mathrm{ConfTS}}= (\widehat{Q}_{1-\alpha}-V(\bm{x},y))^2$$.

\textbf{C-Adapter \citep{liu2024c}}: C-Adapter proposes to optimize an adapter with a Conformal Discriminative (CD) Loss. The CD loss is constructed to improve the $\alpha$-free efficiency of prediction sets:
$\mathcal{L}_{\mathrm{CD}}= \frac{1}{K} \sum_{j\in \mathcal{Y}}\{\sigma(V(\bm{x},y) - V(\bm{x},j))\}$.

\textbf{SCPO \citep{bellotti2021optimized}}: SCPO proposes an approach to train the deep learning model directly by optimizing for maximum predictive efficiency. To enable end-to-end training, the conformal predictor is approximated by a differentiable surrogate objective. Gradient descent is then used to minimize the following loss:
$$\mathcal{L}_{\mathrm{SCPO}}=\sum_{k=1}^K \mathcal{C}_{\theta, k}(\bm{x})+\lambda\left(\mathcal{C}_{\theta, y}(\bm{x})-(1-\alpha)\right)^2,$$where $\lambda>0$ controls the relative importance of the two objective components.

\textbf{Uncertainty-aware (UA) Loss \citep{einbinder2022training}}: Uncertainty-aware Loss introduces a novel training objective that integrates conformal calibration directly into model training by encouraging the conformity scores to follow a uniform distribution. The novel uncertainty-aware component of this loss is:
$$\mathcal{L}_{u}=\sup_{w\in[0,1]}\left|\mathbb{E}[\mathds{1}(V(\boldsymbol{x},y,u)-w]\right|,$$where $V(\boldsymbol{x},y,u) := \sum\limits_{y^\prime \in\mathcal{Y}} \pi_{y^\prime} (\bm{x})\mathds{1}(r_f(\bm{x},y^\prime)< r_f(\bm{x},y))  + u \cdot \pi_y(\bm{x})$.

\subsection{DNN Regressor}

For DNN Regressor, the goal is to predict a continuous target variable \( y \in \mathbb{R} \) based on an input \( \bm{x} \in \mathcal{X} \subset \mathbb{R}^d \), where \( (X, Y) \sim \mathcal{P}_{\mathcal{X}\mathcal{Y}} \) follows a joint data distribution \( \mathcal{P}_{\mathcal{X}\mathcal{Y}} \). Let \( f : \mathcal{X} \to \mathcal{F} \) denote a regression model trained on a dataset \( \{(\bm{x_i}, y_i)\}_{i=1}^n \), where \( \mathcal{F} \) represents the space of model outputs, $\bm{x}_i$ are the features and $y_i$ are the observed target values.

\textbf{ABS \citep{papadopoulos2002inductive}}: 
The ABS score $V(\bm{x}, y) := |f(\bm{x}) - y|$ is designed to measure the absolute deviation of a prediction \(f(\bm{x})\) from the observed target value \(y\), where \(f(\bm{x})\) is a point estimate of given \( \{\bm{x}\} \). 
Then the prediction set for a new instance \(\bm{x}_{n+1}\) is constructed by $\mathcal{C}(\bm{x}_{n+1}) = [f(\bm{x}_{n+1}) - \widehat{Q}_{1-\alpha}, f(\bm{x}_{n+1}) + \widehat{Q}_{1-\alpha}].$

\textbf{NORABS~\citep{papadopoulos2011regression,papadopoulos2011reliable}}: 
The NORABS score $V(\bm{x}, y) := \frac{|f(\bm{x}) - y|}{\gamma + \xi(\bm{x})}$ normalizes the regression error using difficulty measure deviation $\xi(\bm{x})$. The prediction set for $\bm{x}_{n+1}$ is $\mathcal{C}(\bm{x}_{n+1}) = [f(\bm{x}_{n+1}) - \widehat{Q}_{1-\alpha} \cdot(\gamma + \xi(\bm{x}_{n+1})), f(\bm{x}_{n+1}) + \widehat{Q}_{1-\alpha} \cdot(\gamma + \xi(\bm{x}_{n+1}))]$. Additionally, we provide the normalized score for Gaussian Process Regression~\citep{papadopoulos2024guaranteed}.

\textbf{CQR \citep{romano2019conformalized}}: The CQR score \(V(\bm{x}, y) := \max( f(\bm{x})_{\alpha/2} - y, \; y - \)\(f(\bm{x})_{1-\alpha/2} )\) measures the maximum absolute deviation between the true value \( y \) and the predicted quantile interval \([f(\bm{x})_{\alpha/2}, f(\bm{x})_{1-\alpha/2}]\). 
Then the prediction set for a new instance \(\bm{x}_{n+1}\) is constructed by:
$\mathcal{C}(\bm{x}_{n+1}) = [f(\bm{x})_{\alpha/2}  - \widehat{Q}_{1-\alpha},\; f(\bm{x})_{1-\alpha/2} + \widehat{Q}_{1-\alpha}].$

\textbf{CQRR \citep{sesia2020comparison}}: The CQRR score extends the standard CQR method by introducing a scaling mechanism. This score normalizes the deviations between the true value \( y \) and the predicted quantile interval \([f(\bm{x})_{\alpha/2}, f(\bm{x})_{1-\alpha/2}]\) by the width of the interval itself. Formally, for a data pair \((\bm{x}, y)\), the CQRR score is defined as
$
V(\bm{x}, y) := \max\left(\frac{f(\bm{x})_{\alpha/2} - y}{\hat{s}_{\alpha}(\bm{x})}, \; \frac{y - f(\bm{x})_{1-\alpha/2}}{\hat{s}_{\alpha}(\bm{x})} \right),
$
where $\hat{s}_{\alpha}(\bm{x}):=f(\bm{x})_{1-\alpha/2} - f(\bm{x})_{\alpha/2} + \epsilon$ is the scaling factor, and \(\epsilon > 0\) is a small constant added to prevent division by zero. 
The prediction set is given by
$
\mathcal{C}(\bm{x}_{n+1}) = \left[f(\bm{x}_{n+1})_{\alpha/2} - \widehat{Q}_{1-\alpha} \cdot \hat{s}_{\alpha}(\bm{x}_{n+1}), \; f(\bm{x}_{n+1})_{1-\alpha/2} + \widehat{Q}_{1-\alpha} \cdot  \hat{s}_{\alpha}(\bm{x}_{n+1}) \right].$

\textbf{CQRM \citep{sesia2020comparison}}: The CQRM score extends the CQRR score by introducing two scaling factors to independently normalize the deviations for the lower and upper bounds of the prediction interval. This method requires the model to output predictions at three quantile levels: \([\alpha/2, 1/2, 1-\alpha/2]\). 
The two scaling factors are defined as $\hat{s}_{\alpha, \text{lo}}(\bm{x}) := f(\bm{x})_{1/2} - f(\bm{x})_{\alpha/2} + \epsilon$ and $\hat{s}_{\alpha, \text{up}}(\bm{x}) := f(\bm{x})_{1-\alpha/2} - f(\bm{x})_{1/2} + \epsilon$.
The non-conformity score for a data pair \((\bm{x}, y)\) and prediction interval are then defined as:
\[
V(\bm{x}, y) := \max\left(\frac{f(\bm{x})_{\alpha/2} - y}{\hat{s}_{\alpha, \text{lo}}(\bm{x})}, \; \frac{y - f(\bm{x})_{1-\alpha/2}}{\hat{s}_{\alpha, \text{up}}(\bm{x})} \right).
\]
Then, the prediction set is given by 
\[
\mathcal{C}(\bm{x}_{n+1}) = \left[f(\bm{x}_{n+1})_{\alpha/2} - \widehat{Q}_{1-\alpha} \cdot \hat{s}_{\alpha, \text{lo}}(\bm{x}_{n+1}), \; f(\bm{x}_{n+1})_{1-\alpha/2} + \widehat{Q}_{1-\alpha} \cdot \hat{s}_{\alpha, \text{up}}(\bm{x}_{n+1})\right].
\]

\textbf{CQRFM \citep{kivaranovic2020adaptive}}: The CQRFM score is similar to CQRM, as it also requires the model to output predicted quantiles at \([\alpha/2, 1/2, 1-\alpha/2]\). However, it normalizes the deviations from the median \(f(\bm{x})_{1/2}\) by the fractional distance to the nearest quantile bounds. The two fractional scaling factors are defined in the same way as in CQRM.
The non-conformity score and prediction interval are defined as:

\[
V(\bm{x}, y) := \max\left(\frac{f(\bm{x})_{1/2} - y}{\hat{s}_{\alpha, \text{lo}}(\bm{x})}, \; \frac{y - f(\bm{x})_{1/2}}{\hat{s}_{\alpha, \text{up}}(\bm{x})} \right).
\]
Then, the prediction set is given by 
\[
\mathcal{C}(\bm{x}_{n+1}) = \left[f(\bm{x}_{n+1})_{1/2} - \widehat{Q}_{1-\alpha} \cdot \hat{s}_{\alpha, \text{lo}}(\bm{x}_{n+1}), \; f(\bm{x}_{n+1})_{1/2} + \widehat{Q}_{1-\alpha} \cdot \hat{s}_{\alpha, \text{up}}(\bm{x}_{n+1}) \right].
\]

\textbf{R2CCP \citep{guha2023conformal}}: R2CCP converts a regression problem into a classification problem by dividing the range space into \( K \) bins. The model predicts the probability that each sample belongs to one of these \( K \) bins, similar to a standard classification task. To train the regression-to-classification model, R2CCP uses a custom loss function:

\[
\mathcal{L}(\theta) = \sum_{i=1}^{N} \left( \sum_{k=1}^{K} \left( |y_i - m_k|^p \cdot \pi_{ik} \right) - \tau \cdot \sum_{k=1}^{K} \pi_{ik} \log(\pi_{ik}) \right),
\]
where \( \pi_{ik} \) is the predicted probability for the \( k \)-th bin, \( m_k \) are the bin midpoints, \( \tau \) is a regularization parameter, and \( p \) is the exponent applied to the absolute difference between the target value and the bin midpoint. Both \( \tau \) and \( p \) are hyperparameters. After training the regression-to-classification model, R2CCP defines the non-conformity score as the linear interpolation of the model’s softmax probabilities. Based on these scores, R2CCP outputs the interval where all non-conformity scores are smaller than \( \widehat{Q}_{1-\alpha} \).

\textbf{Ensemble CP \citep{xu2021conformal}}: Ensemble CP trains multiple models based on randomly sampled subsets of the data and aggregates their non-conformity scores using a specific aggregation function, such as mean or median. It can update \( \widehat{Q}_{1-\alpha} \) over time, making it adaptable to dynamic time series.

\textbf{ACI \citep{gibbs2021adaptive}}: ACI dynamically adjusts the confidence level $\alpha_t = \alpha_{t-1} + \gamma \left( \alpha - \text{err}_t \right)$ based on the historical error rates to adapt to distribution shifts, where \(\alpha_t\) is the updated confidence level depending on time $t$, \(\alpha\) is the target coverage, and \(\text{err}_t:= \sum_{i=1}^{n} w_i \cdot I(y_{i} \notin \hat{C}_i)\) is the weighted error rate at time step \(t\). This dynamic updating enables ACI to maintain reliable prediction sets despite evolving data distributions.

\textbf{AgACI~\citep{zaffran2022adaptive}}: AgACI is a parameter-free conformal prediction method for time series that builds on ACI by using online expert aggregation. It runs multiple models in parallel with different learning rates $\gamma_k$ and adaptively aggregates their prediction intervals using weighted averages, where weights are based on the historical pinball loss. This approach eliminates the need to tune $\gamma$ manually and ensures valid coverage while improving interval efficiency, especially under temporal dependencies.

\subsection{GNN}

In this section, we discuss the applications of conformal prediction algorithms on graph data. The Graph is defined as $\mathcal{G} = (\mathcal{V}, \mathcal{E})$, where $\mathcal{V}:= \{v_i\}_{i=1}^N$ denotes the node set and $\mathcal{E}$ denotes the edge set with $|\mathcal{E}| = E$. Let $\bm{X} := [\bm{x}_1, \cdot\cdot\cdot, \bm{x}_N ]^T$ be the node feature matrix, where $\bm{x}_i \in \mathbb{R}^d$ is a $d$-dimensional feature vector for node $v_i$. The label of $v_i$ is $y_i \in \mathcal{Y}$, where $\mathcal{Y} := \{1, 2, ..., K\}$ denotes the label space.

\textbf{DAPS \citep{zargarbashi2023conformal}}: DAPS proposes a diffusion-based method that incorporates neighborhood information by leveraging the network homophily. Specifically, the score function is defined as: $
V(\bm{x}_i,y,\lambda) := (1-\lambda) \hat{V}(\bm{x}_i,y) +\frac{\lambda}{\vert\mathcal{N}_i\vert}\sum_{v_j\in\mathcal{N}_i}\hat{V}(\bm{x}_j,y),
$ where $\hat{V}(\cdot, \cdot)$ is the basic non-conformity score function, e.g., APS, and $\lambda$ is a diffusion parameter.

\textbf{SNAPS \citep{song2024similarity}}: SNAPS, which is essentially an enhanced version of the DAPS, aggregates non-conformity scores of nodes with high feature similarity to ego node and one-hop graph structural neighbors. Formally, the score function is shown as:
$
V(\bm{x}_i,y,\lambda, \mu) := (1-\lambda-\mu) \hat{V}(\bm{x}_i,y) +\frac{\lambda}{\bm{D}_s(i,i)}\sum_{j=1}^M\bm{A}_s(i,j)\hat{V}(\bm{x}_j,y)+\frac{\mu}{\vert\mathcal{N}_i\vert}\sum_{v_j\in\mathcal{N}_i}\hat{V}(\bm{x}_j,y),
$
where $\bm{A}_s$ is the adjacency matrix of $k$-NN graph, which measures the similarity between each node pair, and $\bm{D}_s$ is the degree matrix of $\bm{A}_s$. $M$ denotes the number of randomly selected nodes as targets used to construct $k$-NN graph. Both $\lambda$ and $\mu$ are hyperparameters, which are used to measure the importance of three parts of non-conformity scores.

\textbf{NAPS \citep{clarkson2023distribution}}: NAPS, which adapts the weighted variant of CP from \citep{barber2023conformal}, assigns a weight to adjacent nodes, and a weight of zero otherwise. NAPS then applies nonexchangeable conformal prediction with the APS scoring function in the following equation:
$\mathcal{C}(\bm{x}_{n+1}) = \{y\in\mathcal{Y}: V(\bm{x}_{n+1},y)\leq Q_{1-\alpha}\left(\sum_i^n w_i\cdot\delta_{s_i}+w_{n+1}\cdot\delta_{+\infty}\right)\},$
where $Q_{\tau}(\cdot)$ denotes the $\tau$-quantile of a distribution and $s_i=V(\bm{x}_i,y_i)$. Moreover, $w_i$ denotes a weight of $s_i$ and $\delta_{s_i}$ denotes a point mass at $s_i$.

\textbf{CF-GNN \citep{huang2024uncertainty}}: CF-GNN introduces a topology-aware output correction model, akin to Graph Convolutional Network, which employs a conformal-aware inefficiency loss to refine predictions and improve the efficiency of post-hoc CP. For the loss, CF-GNN adopts the optimizing efficiency version of ConfTr.

\subsection{LLM}

\textbf{Conformal LLM \citep{quachconformal}}: This paper proposes a novel conformal procedure based on a risk control framework~\citep{bates2021distribution} for language models by reformulating the generation task as a sampling problem rather than enumerating all possible outputs. The core algorithm samples candidate outputs until a calibrated stopping rule is met, while using a rejection rule to filter out low-quality samples. The method guarantees that the output set contains at least one acceptable answer with high probability while maintaining the statistical guarantees of traditional conformal prediction.

\subsection{Metrics}

In this section, we provide a brief overview of the evaluation metrics supported by \torchcp for diverse modern models, as summarized in Table~\ref{tab:torchcp_evaluation_metrics}. These metrics facilitate rigorous and fine-grained assessment of prediction sets, offering deep insights into model uncertainty.

\begin{table}[!t]
\centering
\small
\begin{tabular}{p{3cm} p{9.6cm}}  
\toprule
\textbf{Deep Models}& \textbf{Metrics} \\
\midrule
\textbf{DNN Classifier} &  Coverage, Size, CovGap~\citep{ding2023class}, WSC~\citep{romano2020classification}, SSCV~\citep{angelopoulos2020uncertainty}, Violated classes~\citep{kasa2023empirically}, DiffViolation~\citep{angelopoulos2020uncertainty}, S Criterion, N Criterion, U Criterion, F Criterion, M Criterion, F Criterion, OU Criterion, OF Criterion, OM Criterion, OE Criterion~\citep{f60917dcc9144362b788a01b03e6894e}, etc.\\
\cmidrule{1-2}
\textbf{DNN Regressor} &  Coverage, Size\\
\cmidrule{1-2}
\textbf{GNN} & 
Coverage, Size, Singleton ratio~\citep{zargarbashi2023conformal} \\
\cmidrule{1-2}
\textbf{LLM} & SetLoss, SSCL~\citep{quachconformal}  \\
\bottomrule
\end{tabular}
\caption{Implemented evaluation metrics in \torchcp.}
\label{tab:torchcp_evaluation_metrics}
\end{table}

\section{Extensibility}
\label{appendix:extensibility}

Each module in \torchcp can be easily extended. In the following, we provide basic guidelines for customizing your own scores, predictors, trainers, and metrics.

\subsection{Add new scores}
\label{extensibility:score}

Adding new scores involves two steps:

\begin{enumerate}
    \item Implementing a New Score Class: new Scores should be implemented in the \texttt{score} module, inheriting from the \texttt{BaseScore} class. Using classification as an example, one could alternatively choose to override the functions: \texttt{\_\_init\_\_()}, \texttt{\_\_call\_\_()}, \texttt{\_calculate\_all\_label()}, and \texttt{\_calculate\_single\_label()}.
    \item Adding an Interface: After customizing the score class, register it in \texttt{score/\_\_init\_\_.py}.
\end{enumerate}

\subsection{Add new predictors}
\label{extensibility:predictor}

Similar to adding new scores, the addition of new predictors also consists of two steps:

\begin{enumerate}
    \item Implementing a New Predictor Class: new Predictors should be implemented in the \texttt{predictor} module, inheriting from the \texttt{BasePredictor} class. Using classification as an example, one could alternatively choose to override the functions: \texttt{\_\_init\_\_()}, \texttt{calibrate()}, and \texttt{predict()}.
    \item Adding an Interface: After customizing the predictor class, register it in \texttt{predictor/\_\_init\_\_.py}.
\end{enumerate}

\subsection{Add new trainers}

Adding new trainers is similar to steps of \ref{extensibility:score} and \ref{extensibility:predictor}:

\begin{enumerate}
    \item Implementing a New Trainer Class: new Trainers should be implemented in the \texttt{trainer} module, inheriting from the \texttt{BaseTrainer} class. Using classification as an example, one could alternatively choose to override the functions: \texttt{\_\_init\_\_()}, \texttt{train()} and \texttt{validate()}. One could pass different loss functions in the \texttt{\_\_init\_\_()} method.
    \item Adding an Interface: After customizing the trainer class, register it in \texttt{trainer/\_\_init\_\_.py}.
\end{enumerate}

\subsection{Add new metrics}

Adding new metrics involves two steps:

\begin{enumerate}
    \item Implementing a New Metric Function: New metrics should be implemented in the \texttt{utils/metrics.py}. 
    \item Adding an Interface: After customizing the metric function, register it by \texttt{Registry(``METRICS'')} in the \texttt{utils/metrics.py}.
\end{enumerate}

\section{Example Codes for \torchcp}
\label{appendix:exmaple_code}
\torchcp provides a collection of example codes in the ``examples'' folder~\footnote{\url{https://github.com/ml-stat-Sustech/TorchCP/tree/master/examples}} of the official codebase, designed to help users quickly get started with various conformal prediction tasks. In this section, we briefly introduce representative examples for the four supported tasks.

\subsection{Example codes for offline conformal prediction}
This part provides brief descriptions of example codes under the offline setting:
\begin{enumerate}
    \item \textbf{DNN Classifier.} This example code\footnote{\url{https://github.com/ml-stat-Sustech/TorchCP/blob/master/examples/classification_splitcp_cifar100.py}} applies Split CP with the LAC score on CIFAR-100 using a pretrained ResNet-20. It evaluates the coverage and average set size of the resulting prediction sets.
    \item \textbf{DNN Regressor.} The example code of regression\footnote{\url{https://github.com/ml-stat-Sustech/TorchCP/blob/master/examples/regression\_cqr\_synthetic.py}} shows the complete workflow of applying Split CP with the Conformalized Quantile Regression score \citep{romano2019conformalized} on a synthetic dataset in a regression task, including model training, calibration, prediction interval generation, and evaluation.
    \item \textbf{GNN.} The GNN example\footnote{\url{https://github.com/ml-stat-Sustech/TorchCP/blob/master/examples/gnn_transductive_coraml.py}} implements a Graph Convolutional Network (GCN)~\citep{DBLP:conf/iclr/KipfW17/GCN} for node classification on the CoraML dataset~\citep{mccallum2000automating} under a transductive setting. Conformal prediction is then applied to generate prediction sets with guaranteed coverage.
    \item \textbf{LLM.} The LLM example\footnote{\url{https://github.com/ml-stat-Sustech/TorchCP/blob/master/examples/llm_ConformalLM_TriviaQA.py}} demonstrates how to deploy conformal prediction in large language model generation tasks, i.e., TriviaQA~\citep{joshi2017triviaqa}.

\end{enumerate}

\subsection{Example codes for semi-online conformal prediction}
\label{appendix:online_cp}

Semi-online conformal prediction means that after predicting a test point, its true label is immediately revealed, and the test sample is incorporated into the calibration set. \torchcp supports this setting as well, with the following examples:

\begin{enumerate}
    \item \textbf{DNN Regressor.} In \torchcp, the semi-online setup of the regression task is integrated into \texttt{ACIPredictor}~\citep{gibbs2021adaptive}. Therefore, we added an example code~\footnote{\url{https://github.com/ml-stat-Sustech/TorchCP/blob/master/examples/timeseries_aci_synthetic.py}} for the regression task in \torchcp. 
    \item \textbf{DNN Classifier.} For DNN classification tasks, this semi-online calibration approach can be naturally implemented in \torchcp by iteratively invoking the ``calibrate()'' function with new incoming data. To clarify this, we have added a semi-online conformal prediction example~\footnote{ \url{https://github.com/ml-stat-Sustech/TorchCP/blob/master/examples/classification_splitcp_cifar100_online.py}} for a classification task in \torchcp.
\end{enumerate}

\section{Empirical Results}
\label{appendix:results}
This section presents empirical results of \torchcp. We first reproduce several classical conformal prediction methods across various tasks. Then, we benchmark the computational efficiency of \torchcp against several representative conformal prediction libraries.

\subsection{Evaluation results}
\begin{figure}[!t]
\centering
  \begin{subfigure}{0.45\textwidth}
    \centering
    \includegraphics[width=\linewidth]{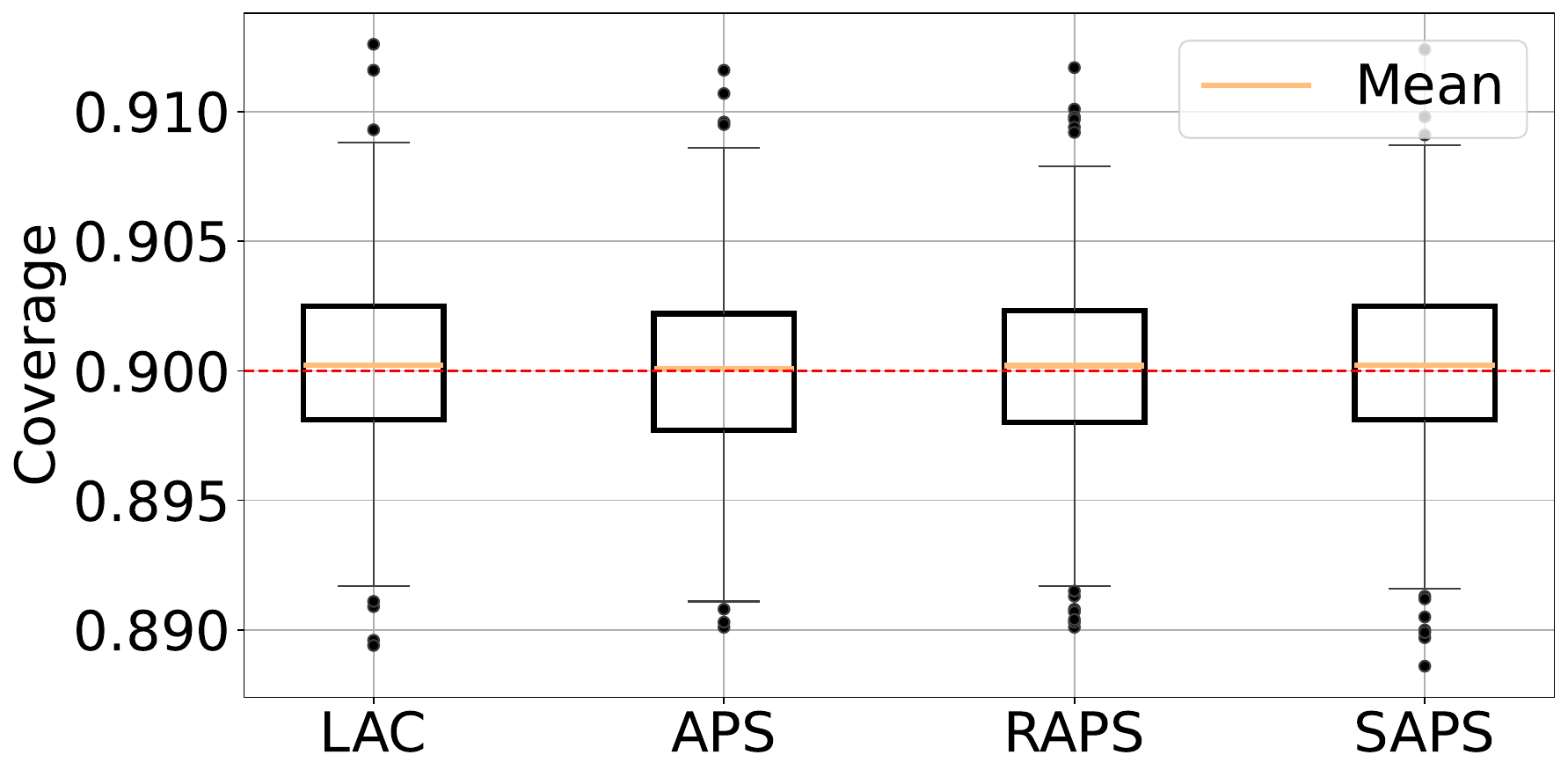}
    \caption{Coverage}
  \end{subfigure}
  \begin{subfigure}{0.45\textwidth}
    \centering
    \includegraphics[width=\linewidth]{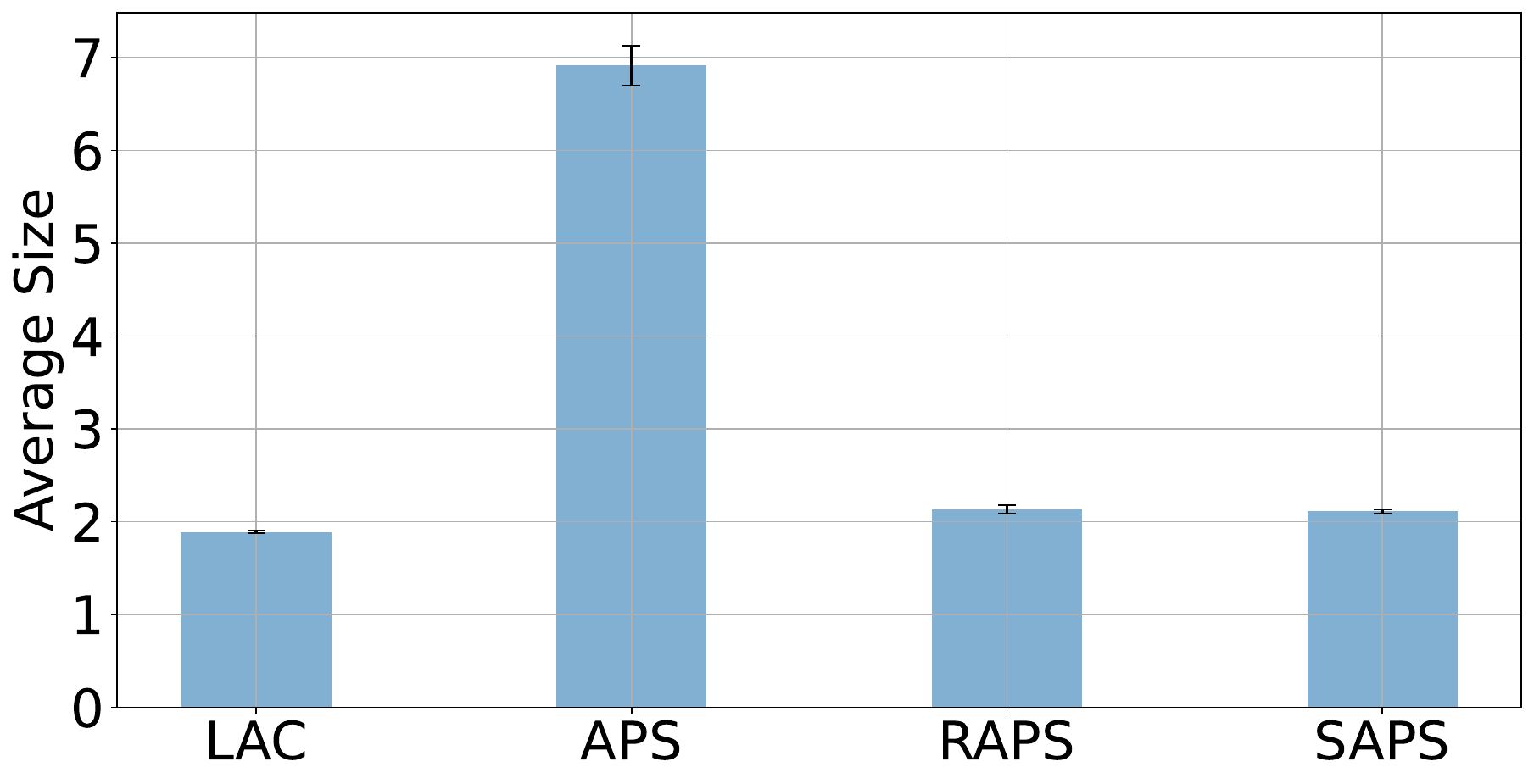}
    \caption{Size}
  \end{subfigure}
  \caption{Results of image classification with $\alpha=0.1$ on ImageNet dataset.}
  \label{fig:calssification_split}
\end{figure}

\begin{figure}[!htbp]
\centering
  \begin{subfigure}{0.3\textwidth}
    \centering
    \includegraphics[width=\linewidth]{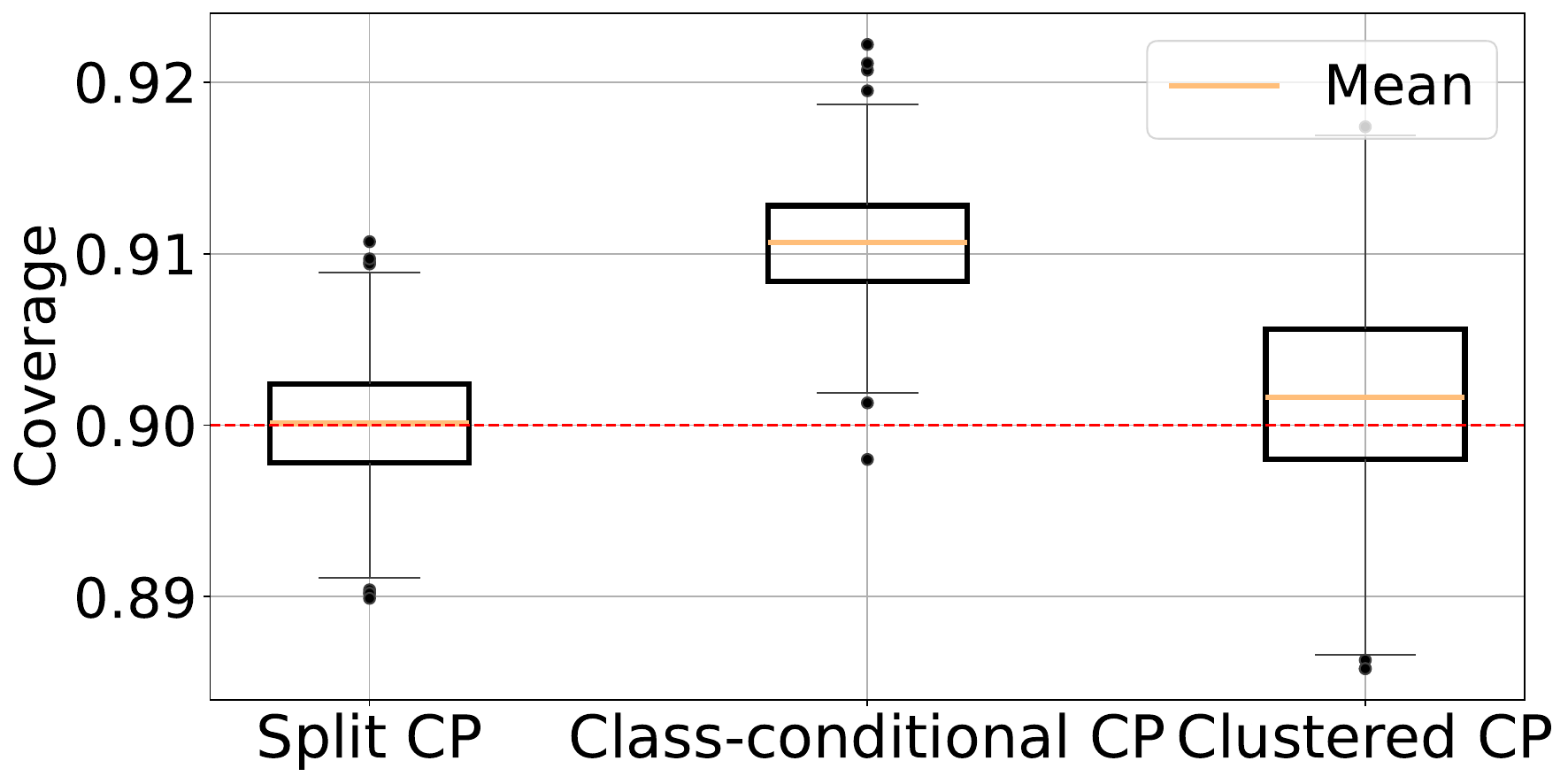}
    \caption{Coverage}
  \end{subfigure}
  \begin{subfigure}{0.3\textwidth}
    \centering
    \includegraphics[width=\linewidth]{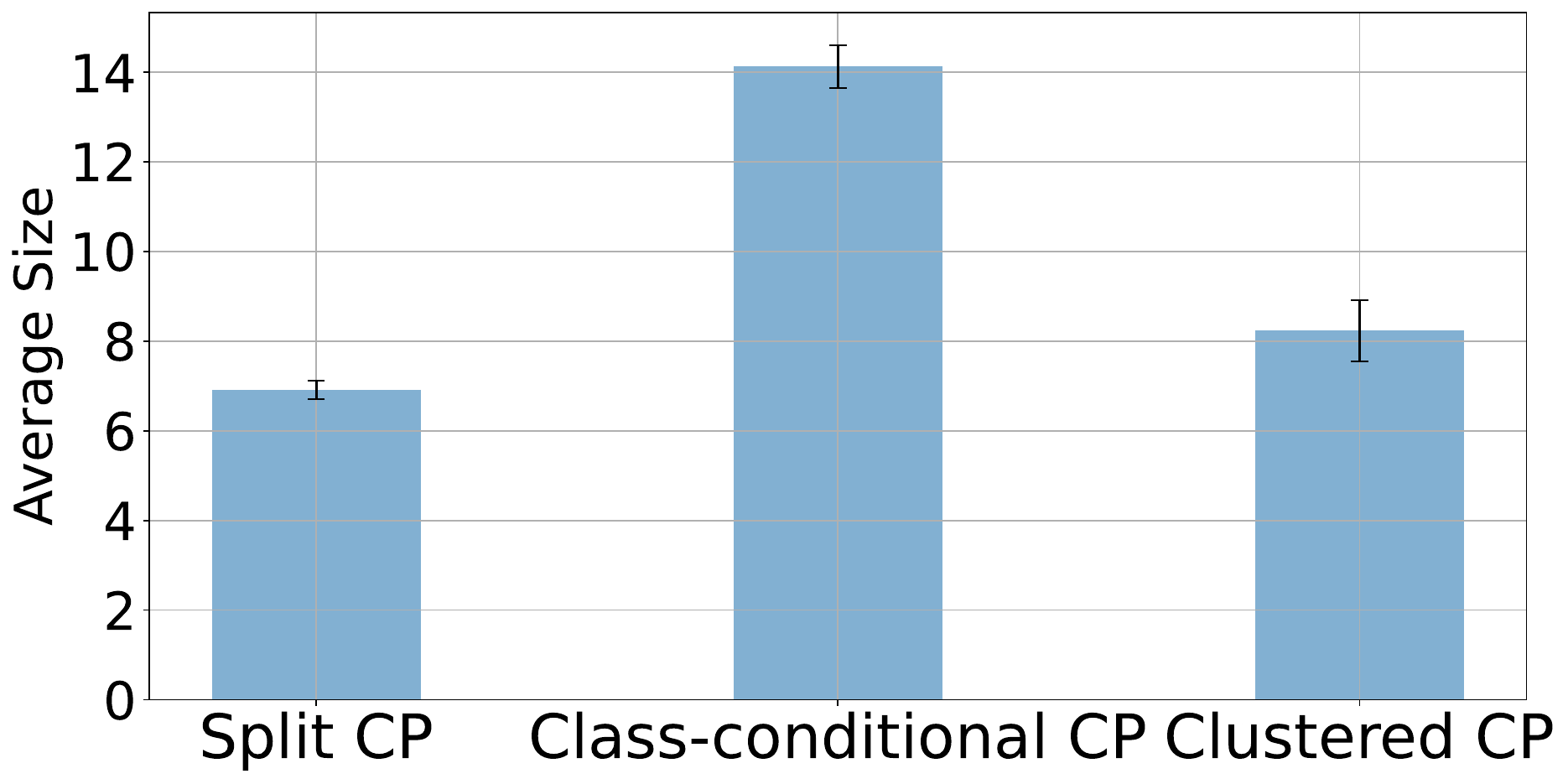}
    \caption{Size}
  \end{subfigure}
  \begin{subfigure}{0.3\textwidth}
    \centering
    \includegraphics[width=\linewidth]{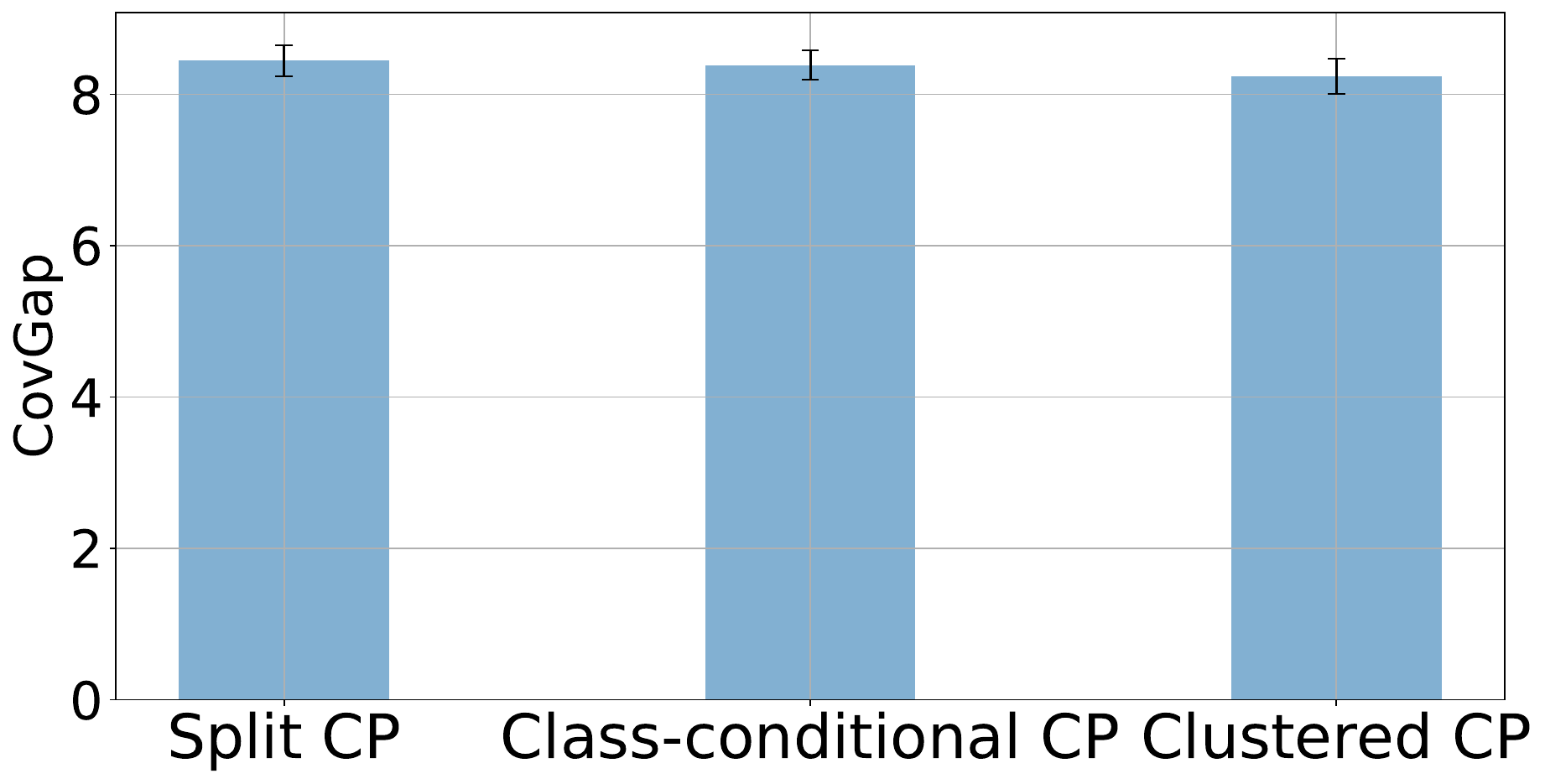}
    \caption{CovGap}
  \end{subfigure}
  \caption{Conditional coverage results on ImageNet image classification with $\alpha = 0.1$.}
  \label{fig:calssification_class}
\end{figure}

\paragraph{DNN Classifier.} In Figure~\ref{fig:calssification_split}, we present the experimental results of Split CP on ImageNet, using LAC, APS, RAPS ($\lambda$=0.1, $k$ =0), and SAPS ($\lambda$=0.2). 
Figure~\ref{fig:calssification_class} compares Split CP, Class-conditional CP, and Clustered CP on the ImageNet dataset, each employing the APS scoring rule. Results are averaged over 1,000 independent trials.

\begin{figure}[!t]
\centering
  \begin{subfigure}{0.45\textwidth}
    \centering
    \includegraphics[width=\linewidth]{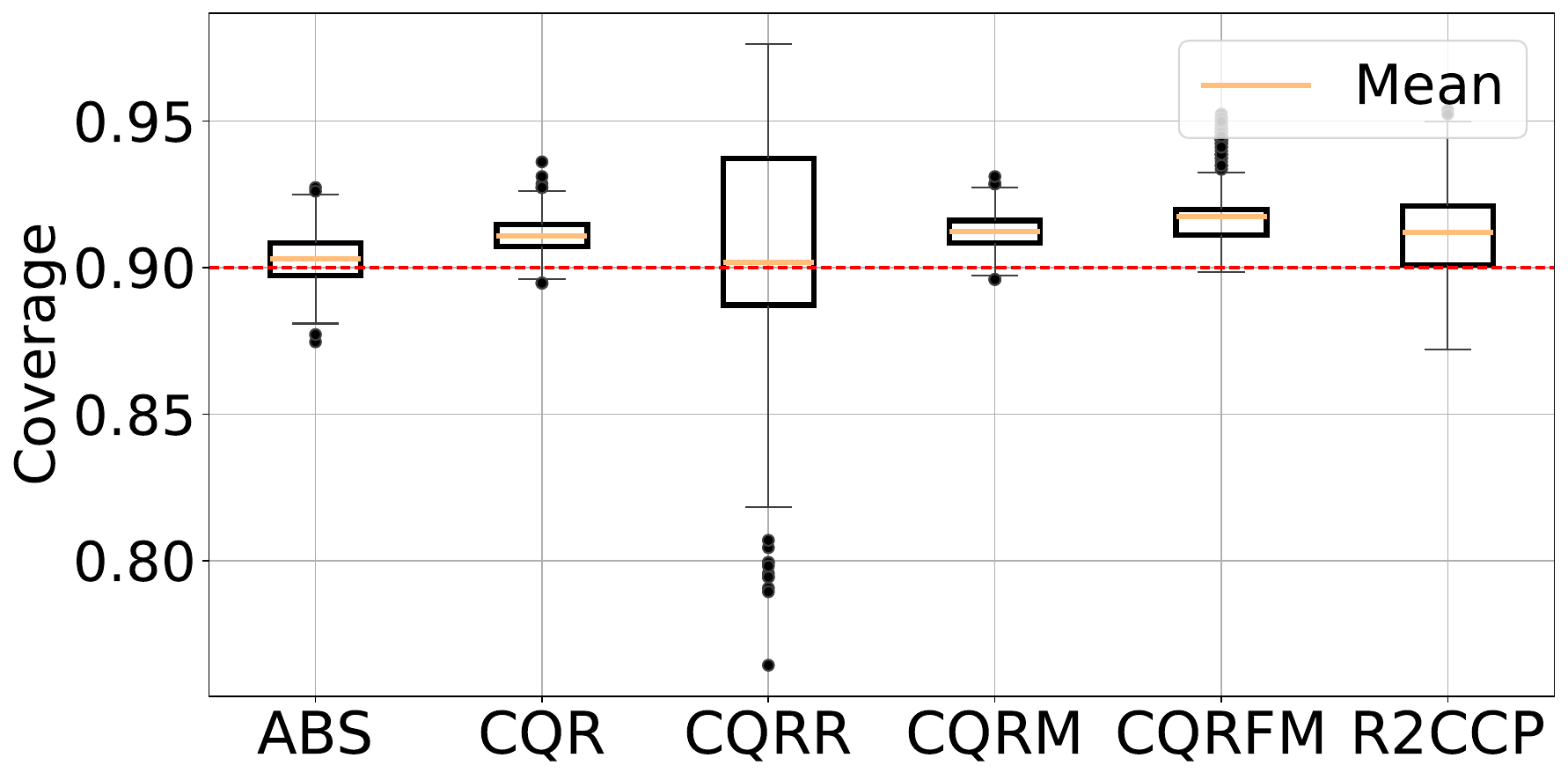}
    \caption{Coverage}
  \end{subfigure}
  \begin{subfigure}{0.45\textwidth}
    \centering
    \includegraphics[width=\linewidth]{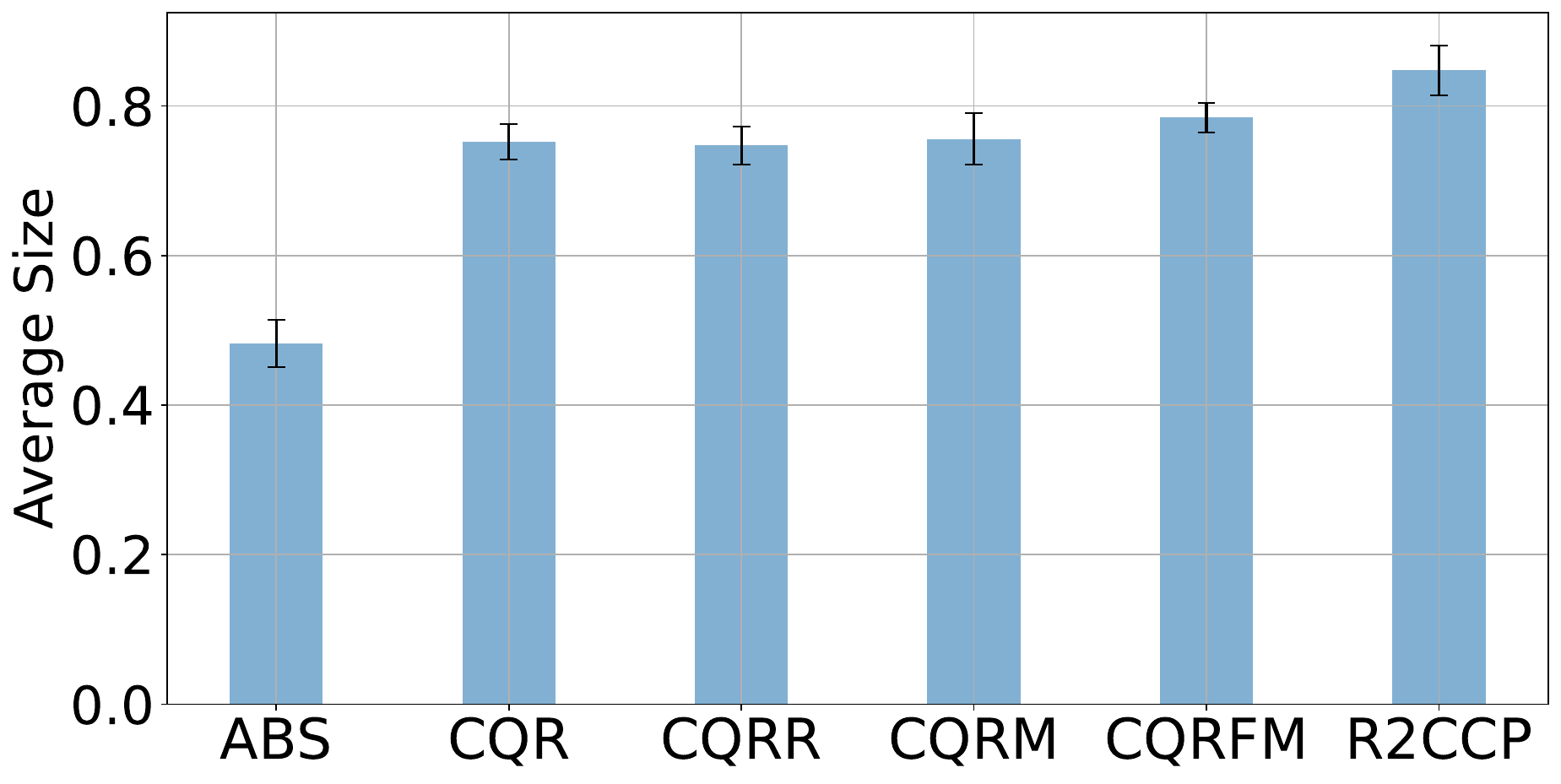}
    \caption{Size}
  \end{subfigure}
  \caption{Results of regression with $\alpha=0.1$ on Community dataset.}
  \label{fig:reg_split}
\end{figure}

\vspace{-10pt}
\paragraph{DNN Regressor.}
In Figure~\ref{fig:reg_split}, we present an example of experimental results on the Community dataset \citep{communities_and_crime_183}, using ABS, CQR, CQRR, CQRM, CQRFM, R2CCP with Split CP. The number of trials is 1000.

\begin{figure}[!t]
\centering
  \begin{subfigure}{0.45\textwidth}
    \centering
    \includegraphics[width=\linewidth]{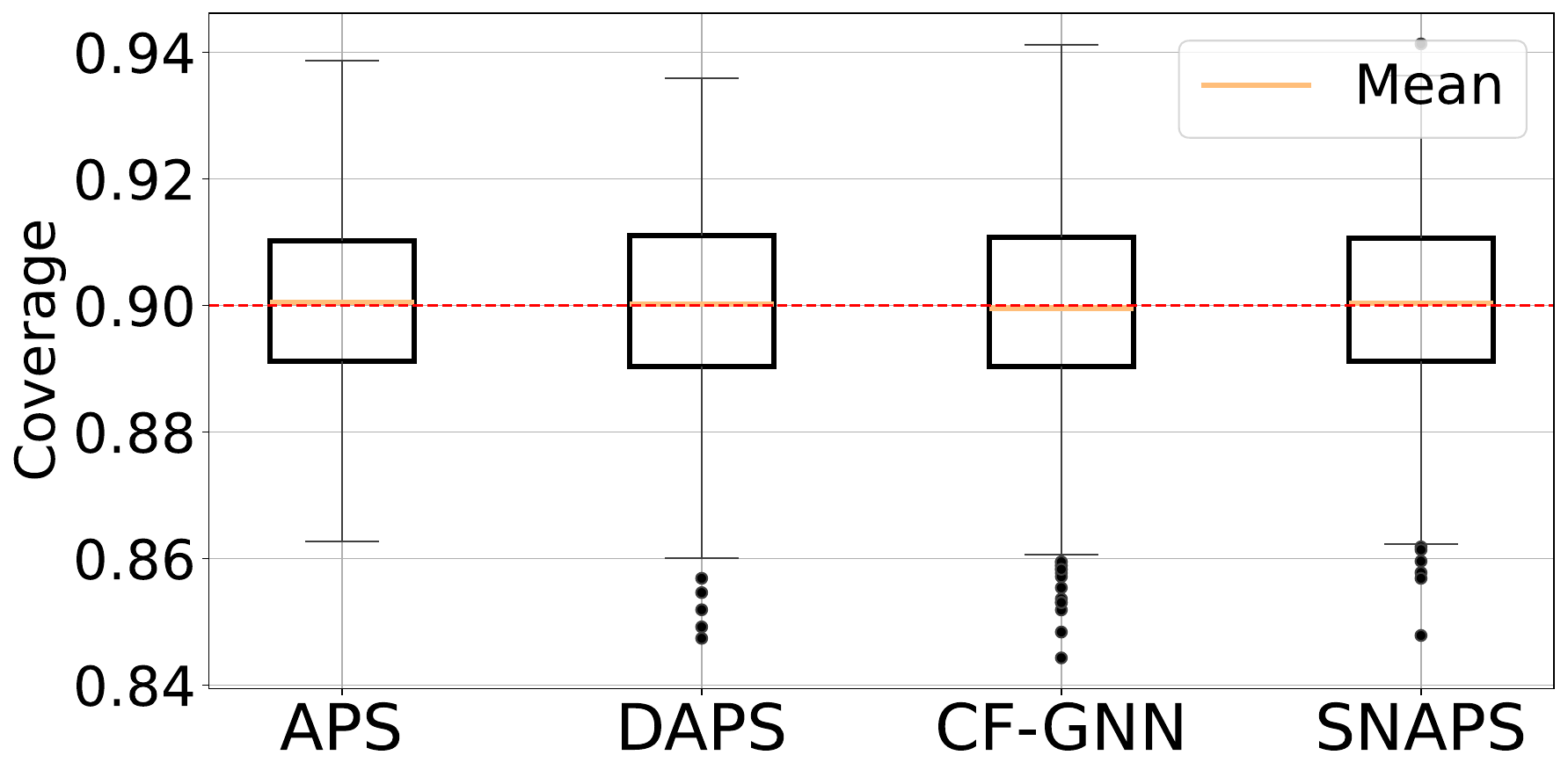}
    \caption{Coverage}
  \end{subfigure}
  \begin{subfigure}{0.45\textwidth}
    \centering
    \includegraphics[width=\linewidth]{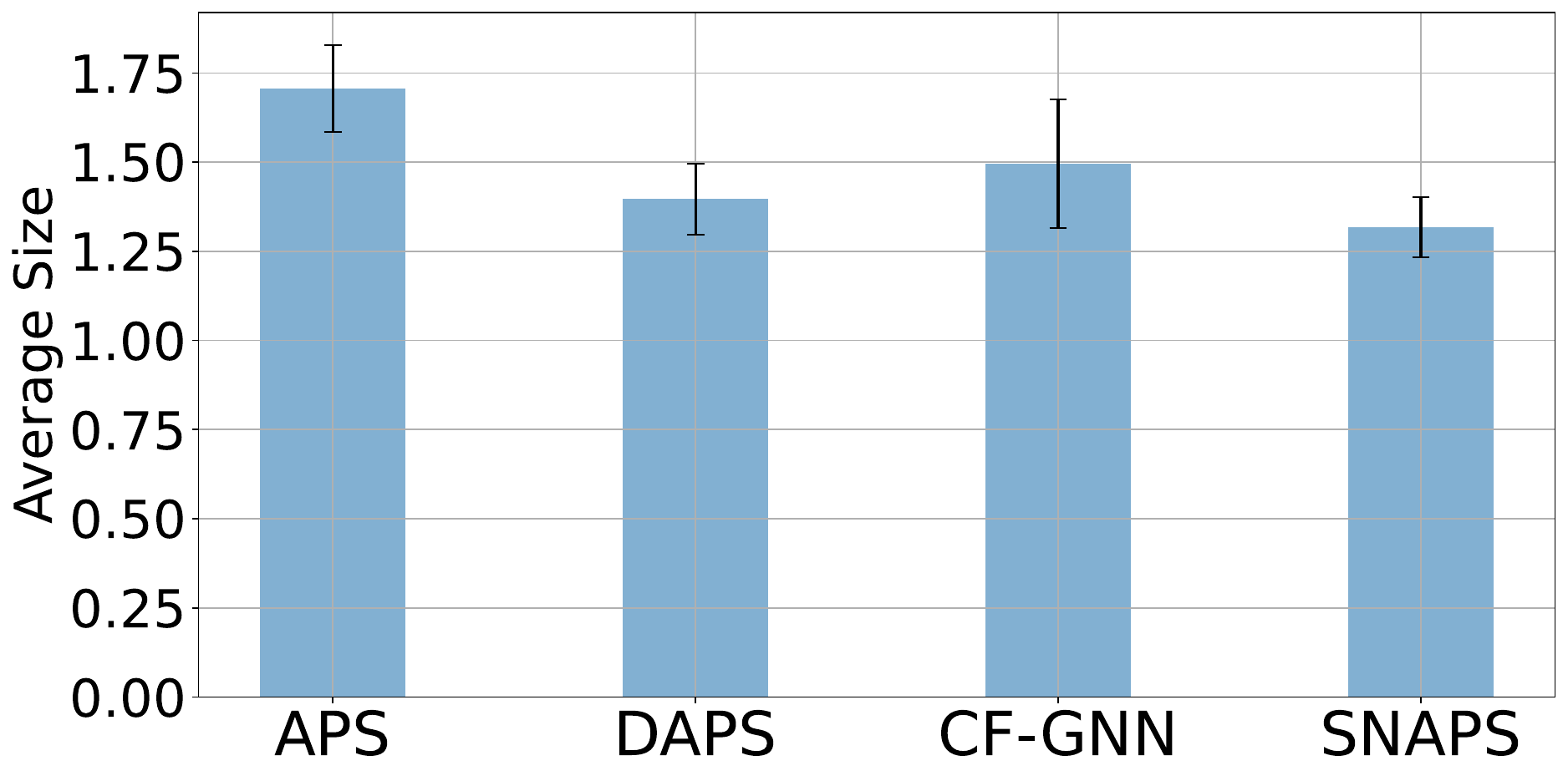}
    \caption{Size}
  \end{subfigure}
  \caption{Results of Graph node classification with $\alpha=0.1$ on CoraML dataset.}
  \label{fig:graph}
\end{figure}

\vspace{-5pt}
\paragraph{GNN.}
In Figure~\ref{fig:graph}, we present experimental results on the CoraML dataset \citep{mccallum2000automating} for node classification under a transductive learning setting.
 We evaluate four methods: APS, DAPS, CF-GNN, and SNAPS, each conducted over 1000 independent trials. For DAPS, the diffusion parameter is set to 0.5. In CF-GNN, the topology-aware correction model is implemented using a two-layer GCN. For SNAPS, the hyperparameters are set as follows: $\lambda = \frac{1}{3}$, $\mu = \frac{1}{3}$, and $k = 20$. These hyperparameter choices follow the default or recommended configurations reported in the respective original papers.

\paragraph{LLM.} 
In Table~\ref{table:llm}, we present an example prediction set, generated by conformal LLM, of the TriviaQA dataset~\citep{joshi2017triviaqa}. The generation process terminates after the fifth sample, as the cumulative score surpasses the conformal threshold.

\begin{table}[!t]
    \centering
    \begin{tabular}{|c|c|c|c|}
        \hline
        Question &   \multicolumn{3}{c|}{\centering Men Against the Sea and Pitcairn's Island were two sequels to what famous novel?} \\
        \hline
         & Answer & Score & Label \\
        \hline
        Answer 1 & Mutiny On The Bounty &0.5347 & True\\
        \hline
        Answer 2 & Treasure Island & 0.5087 & False\\
        \hline
        Answer 3 & Treasure Island & 0.5087 & False\\
        \hline
        Answer 4 & Mutiny on the Bounty & 0.1889 & True\\
        \hline
        Answer 5 & Robinson Crusoe & 0.1384 & False\\
        \hline
    \end{tabular}
    \caption{An Example of prediction sets from Conformal LLM on TriviaQA.}
    \label{table:llm}
\end{table}

\subsection{Comparison of computational efficiency}
\label{appendix:efficiency}

\subsubsection{Classification Task}

We evaluated the computational efficiency of \torchcp in image classification tasks by comparing its calibration and inference time with several existing CP libraries. Specifically, we use two standard benchmarks: MNIST~\citep{lecun2010mnist} and ImageNet~\citep{imagenet15russakovsky}. Throughout all experiments, we used Split CP with APS as the conformal prediction method.
Moreover, \torchcp supports two data processing modes: full-batch (processing the entire dataset at once), and batch (processing data in mini-batches). In our experiments, we set the batch size to 128 for batch processing. We use NVIDIA H100 for GPU computation and AMD EPYC 9654 for CPU computation.

\begin{figure}[!t]
\centering
  \begin{subfigure}{0.45\textwidth}
    \centering
    \includegraphics[width=\linewidth]{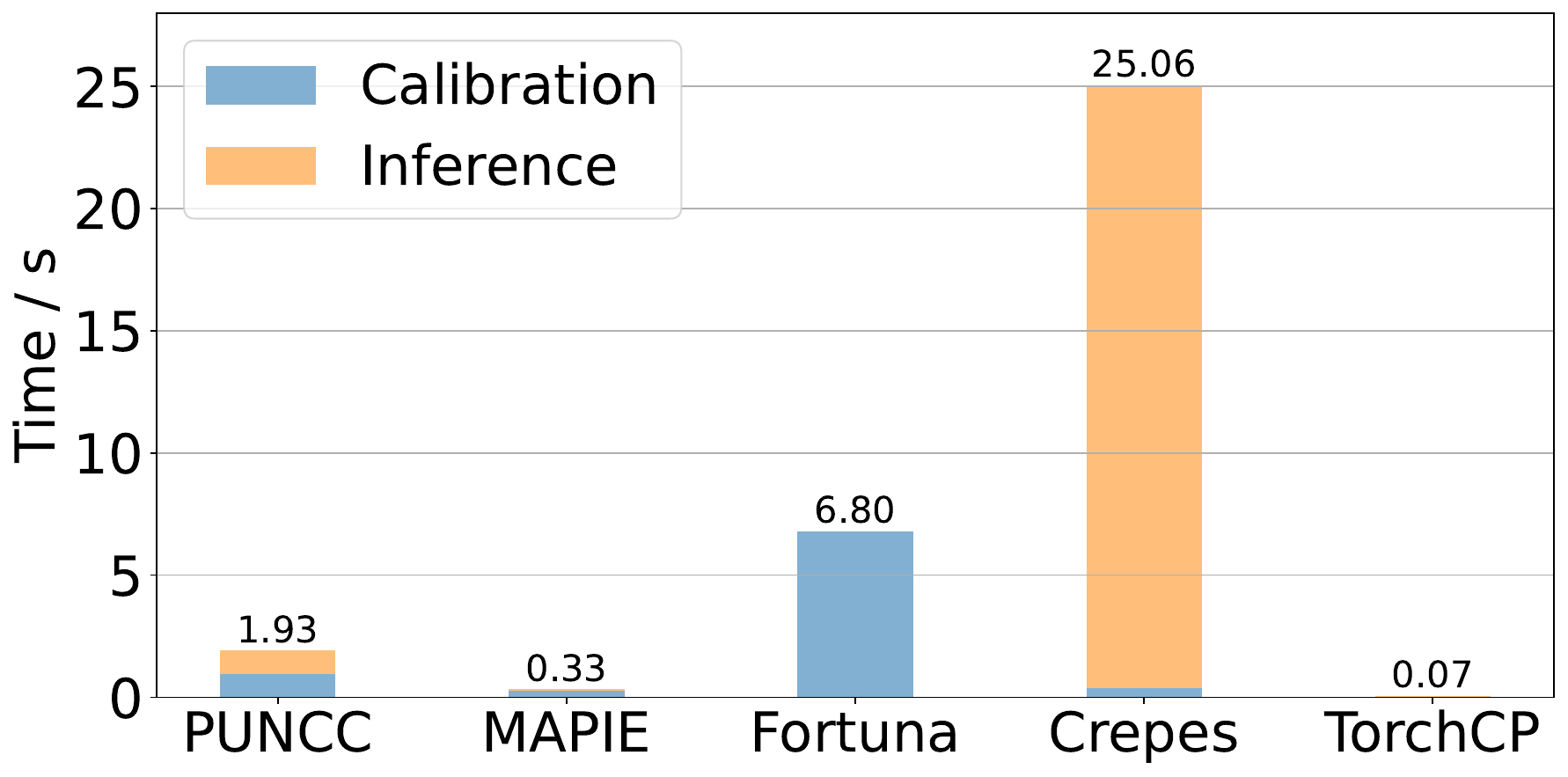}
    \caption{MNIST}
    \label{fig:minst_consuming_time_diff}
  \end{subfigure}
  \begin{subfigure}{0.45\textwidth}
    \centering
    \includegraphics[width=\linewidth]{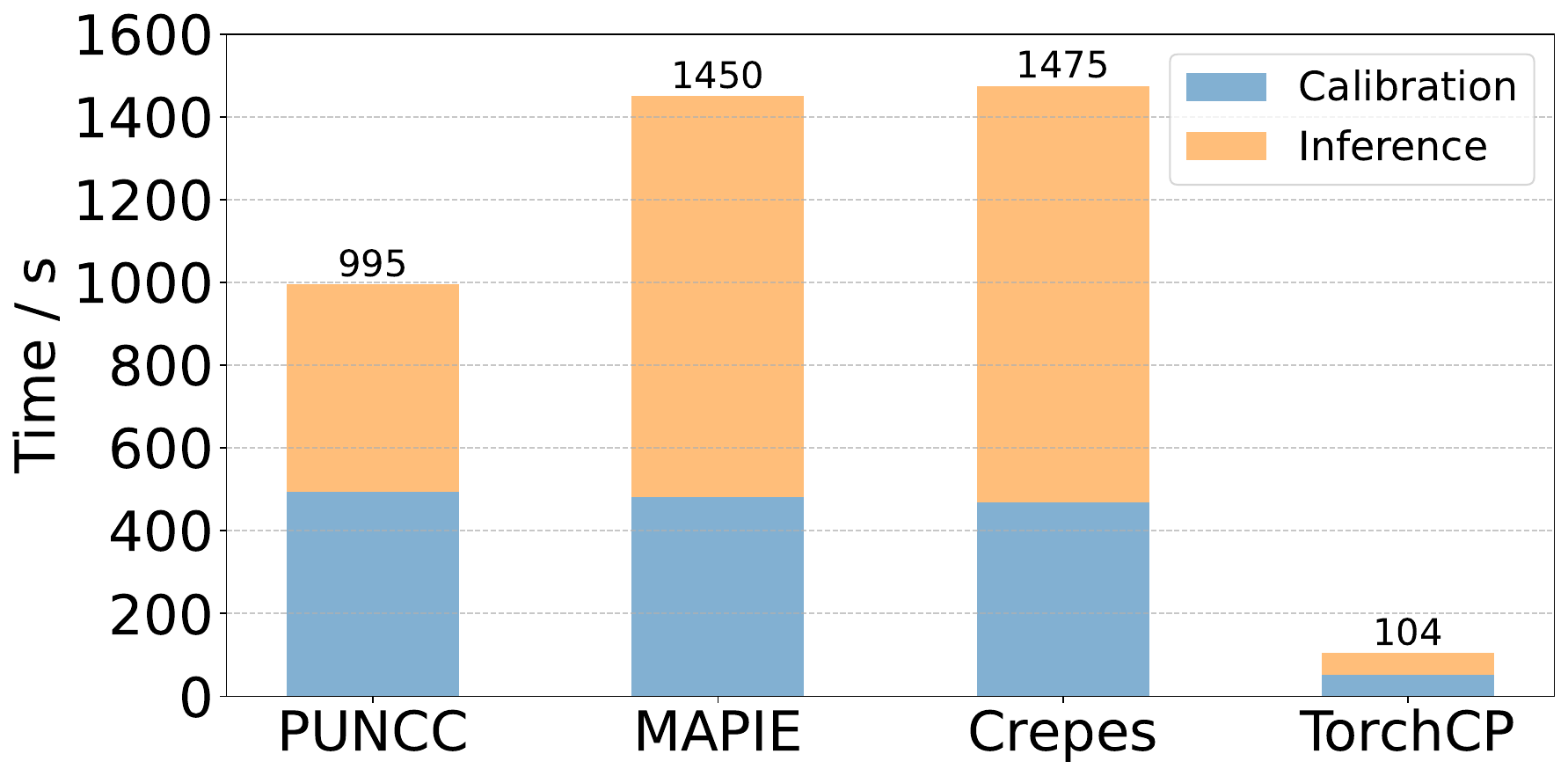}
    \caption{ImageNet}
    \label{fig:imagenet_consuming_time_diff}
  \end{subfigure}
  \caption{Runtime comparison of different conformal prediction libraries.}
\end{figure}

\textbf{MNIST.} We compared the computational efficiency of five libraries: \texttt{PUNCC}, \texttt{MAPIE}, \texttt{Fortuna}, \texttt{Crepes}, and \torchcp (GPU-based full-batch processing). For the dataset, we use 48,000 samples for training, 6,000 for calibration, and 6,000 for testing. For \torchcp, we use a LeNet-5 model for training and inference using PyTorch on the GPU. For \texttt{PUNCC}, \texttt{MAPIE}, and \texttt{Crepes}, we obtain model predictions using PyTorch on a GPU, and then convert output probabilities to NumPy arrays for conformal calibration and inference using CPU. 
For \texttt{Fortuna}, we train the LeNet-5 model using the built-in training algorithm and perform model predictions with JAX on a GPU. We only report the total runtime for \texttt{Fortuna} as it ties the conformal calibration and inference processes together. Figure~\ref{fig:minst_consuming_time_diff} presents the running times of the five libraries on the MINST dataset.  
The results show that \torchcp achieves the lowest calibration and inference time among the five libraries. 
Specifically, TorchCP completes calibration and inference for the entire dataset in just 0.07 seconds, using only 3.6\% of the runtime needed by \texttt{PUNCC} (1.93 seconds).
Moreover, the \texttt{Crepes} library exhibits the longest inference time on the MNIST dataset, likely due to its computation of p-values for each test sample's non-conformity scores by comparing them against the entire calibration set.
This substantial improvement highlights the efficiency advantage of \torchcp on small-scale datasets.

\textbf{ImageNet.} We compared the computational efficiency of four libraries: \texttt{PUNCC}, \texttt{MAPIE}, \texttt{Crepes}, and \torchcp (GPU-based batch processing). In the dataset, we use a pre-trained ResNet-50 model from \texttt{torchvision} and select 25,000 images from the validation set for calibration and another 25,000 for testing. For \torchcp, we leverage GPU-based batch processing to efficiently run the conformal pipeline. For \texttt{PUNCC}, \texttt{MAPIE}, and \texttt{Crepes}, we conducted experiments on the CPU, as they lack batch processing support, and their GPU-based full-batch processing causes out-of-memory errors.
Figure~\ref{fig:imagenet_consuming_time_diff} shows the results of running times of three libraries on the ImageNet dataset.  
The results demonstrate that TorchCP achieves the shortest calibration and inference times, cutting approximately 90\% of the time required by \texttt{PUNCC} (104s vs. 995s). 
This significant enhancement underscores the scalability of TorchCP on large-scale datasets and models.

\begin{figure}[htbp]
\centering
  \begin{subfigure}{0.45\textwidth}
    \centering
    \includegraphics[width=\linewidth]{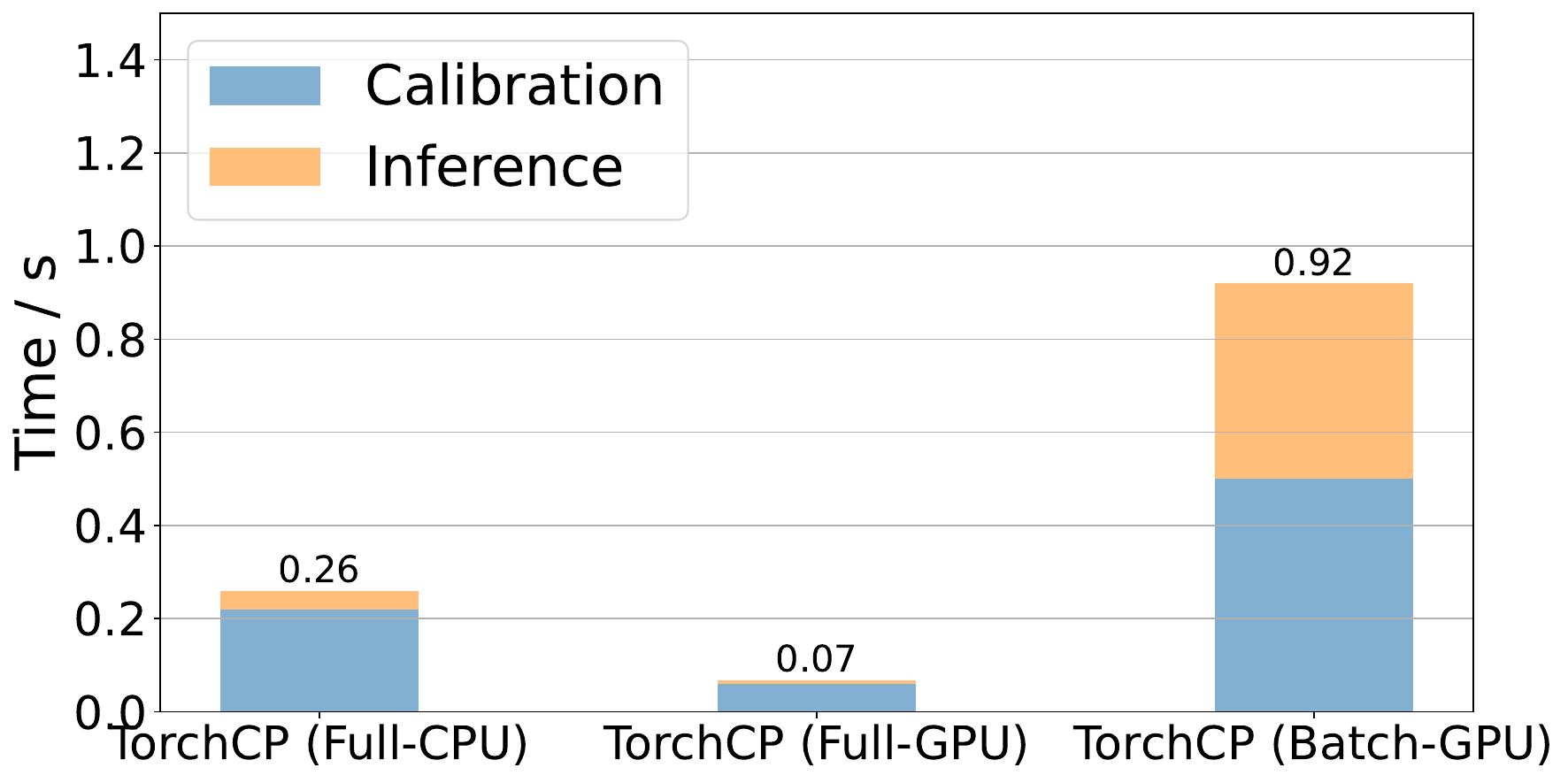}
    \caption{MNIST}
    \label{fig:consuming_time_torchcp_minst}
  \end{subfigure}
  \begin{subfigure}{0.45\textwidth}
    \centering
    \includegraphics[width=\linewidth]{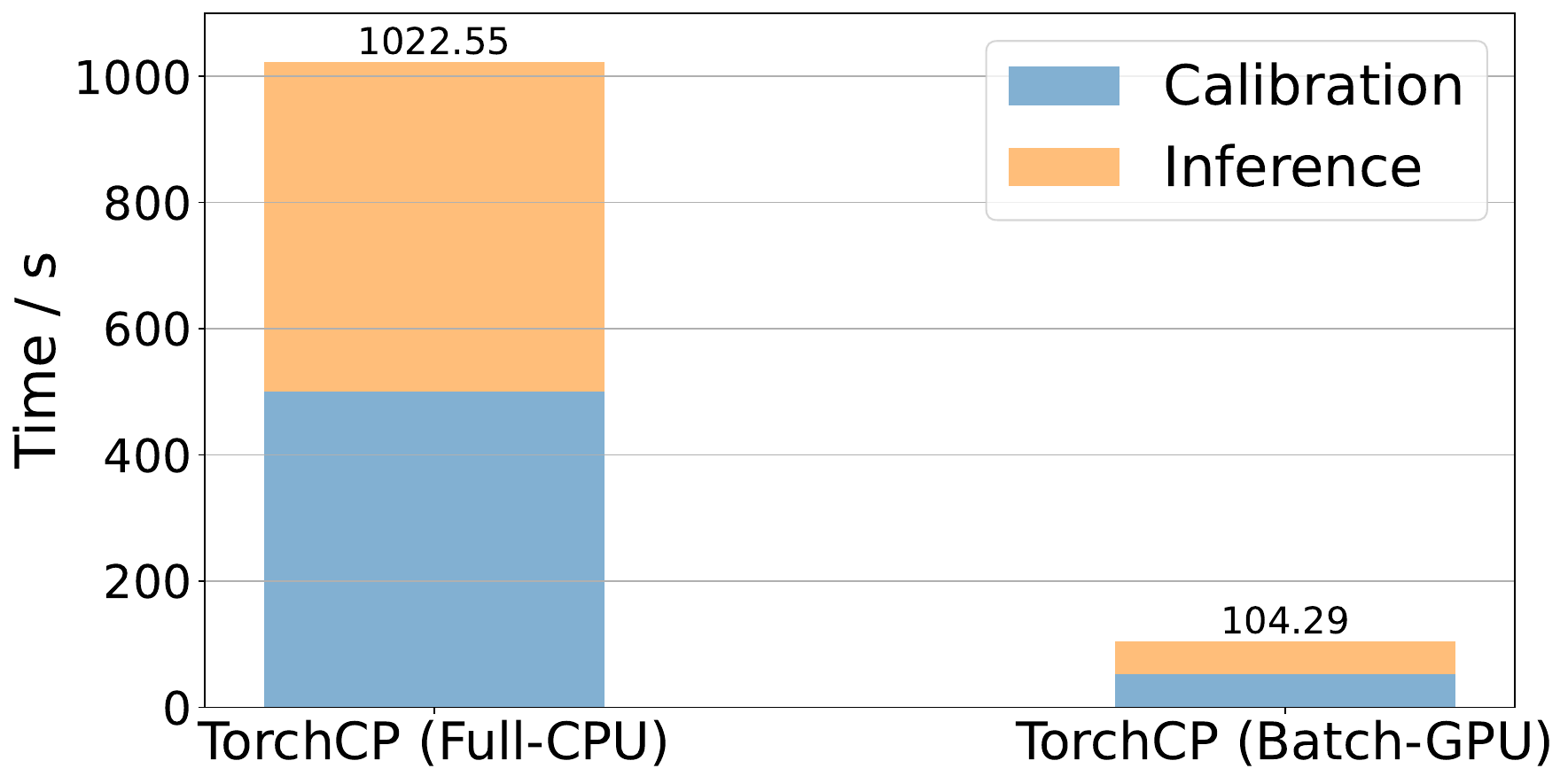}
    \caption{ImageNet}
    \label{fig:consuming_time_torchcp_imagenet}
  \end{subfigure}
  \caption{Runtime comparison of different data processing modes.}
\end{figure}

\textbf{Choices of batch processing.} To assess the efficiency of two processing modes (Full-batch and batch), we conduct experiments with \torchcp on both MNIST and ImageNet, using the same data splits and models as in prior experiments. 
We utilize GPU-accelerated computing for all experiments except the full-batch mode on ImageNet, where the dataset's size exceeds GPU memory capacity.
Figures~\ref{fig:consuming_time_torchcp_minst} and ~\ref{fig:consuming_time_torchcp_imagenet} present the running times of different data processing modes on MNIST and ImageNet, respectively. 
On the small-scale dataset, i.e., MNIST, using full-batch processing on GPU achieves the fastest implementation among the three choices, requiring only 26\% computational time of that with CPU.
When it comes to the ImageNet dataset, GPU-based batch processing completes calibration and inference in approximately 104.29 seconds, compared to 1022.55 seconds for CPU-based processing, achieving a 9.8$\times$ speedup. These findings suggest that GPU-based full-batch processing is optimal for small-scale datasets where GPU memory is sufficient to load the whole dataset. Conversely, GPU-based batch processing is preferable for large-scale datasets like ImageNet, offering significant runtime advantages and enabling scalable, efficient deployment.

\begin{figure}[htbp]
\centering
  \begin{subfigure}{0.22\textwidth}
    \centering
    \includegraphics[width=\linewidth]{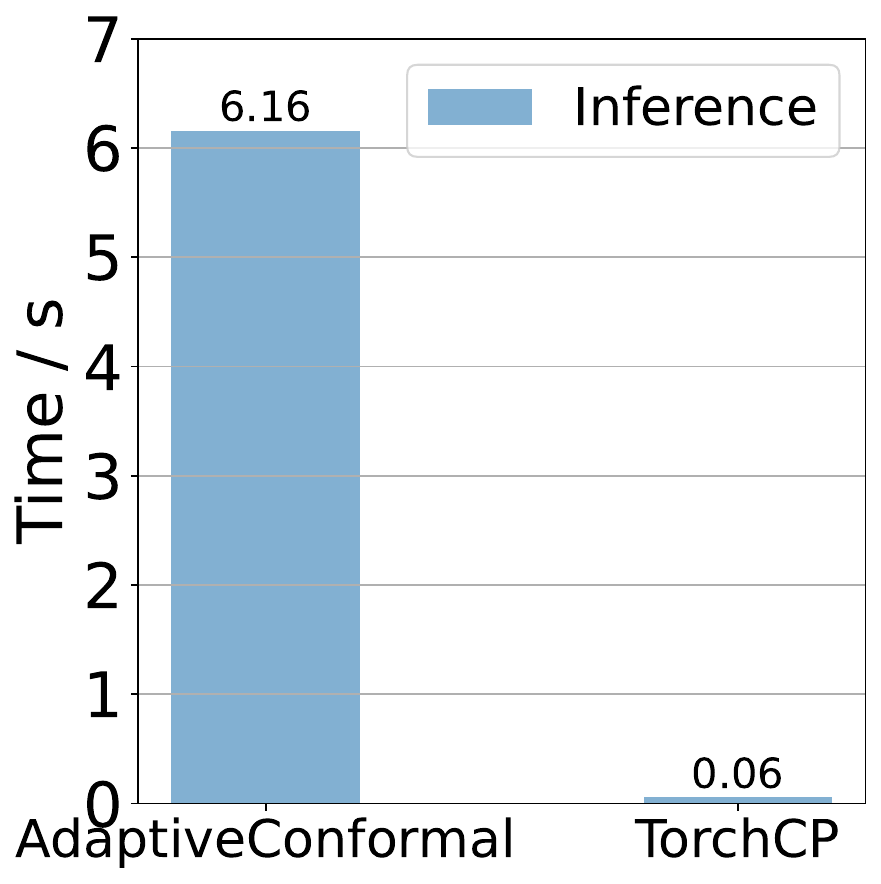}
    \caption{Time}
    \label{fig:consuming_time_regression}
  \end{subfigure}
  \begin{subfigure}{0.36\textwidth}
    \centering
    \includegraphics[width=\linewidth]{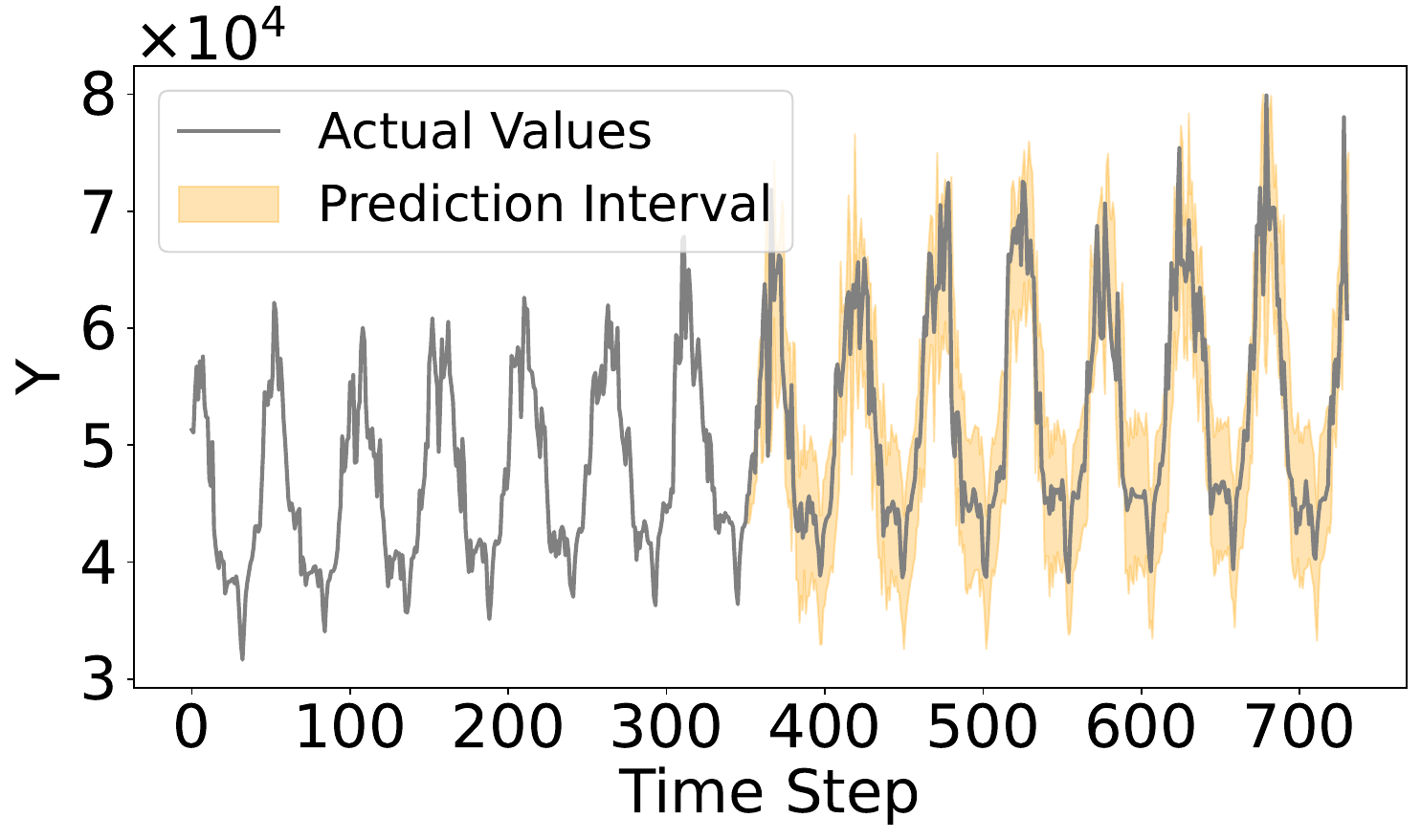}
    \caption{AdaptiveConformal}
    \label{fig:AdConformal_prediction}
  \end{subfigure}
  \begin{subfigure}{0.36\textwidth}
    \centering
    \includegraphics[width=\linewidth]{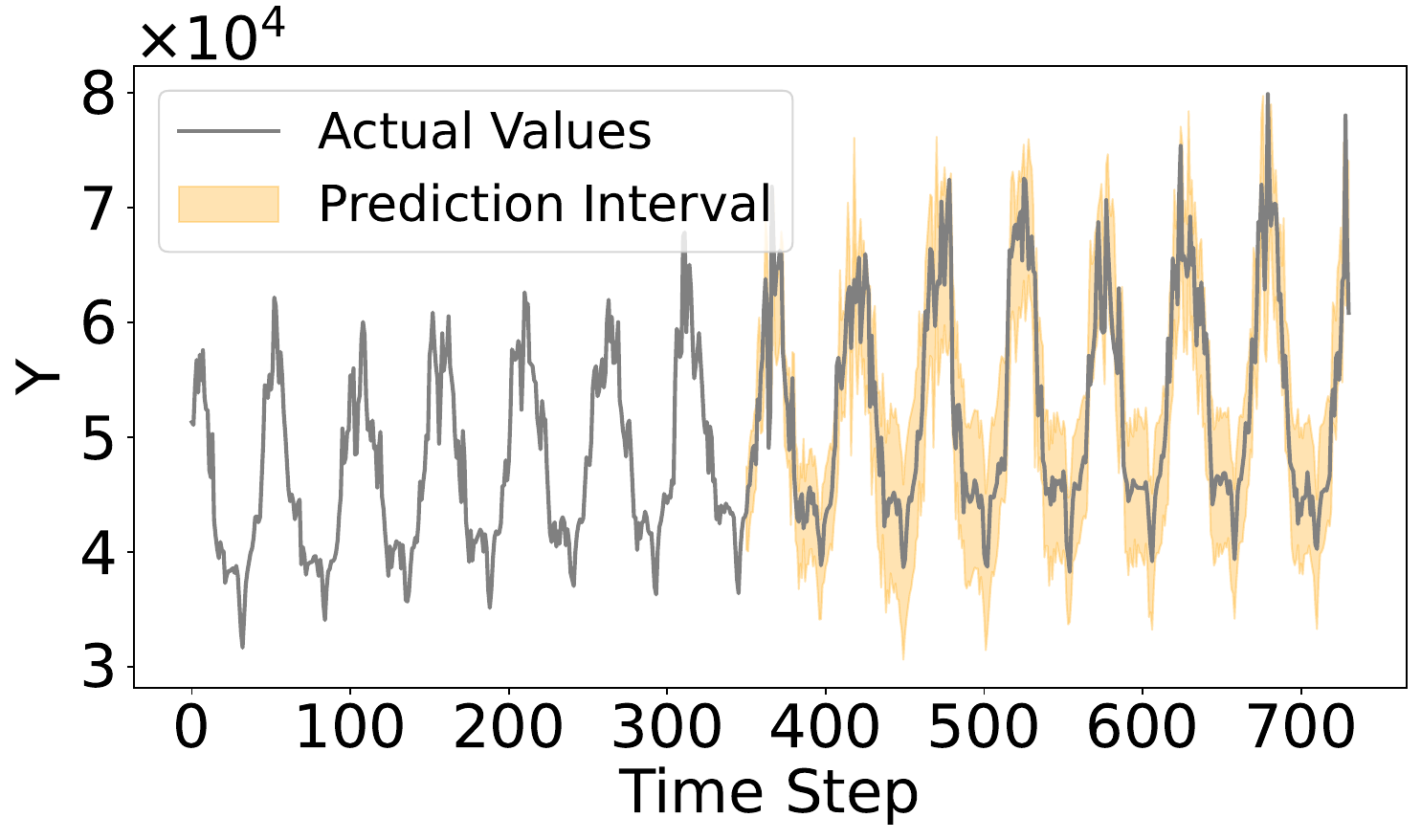}
    \caption{TorchCP}
    \label{fig:torchcp_prediction}
  \end{subfigure}
  \caption{Comparison of time consumption and prediction intervals between AdaptiveConformal and TorchCP on the France electricity dataset.}
  \label{fig:conusming_time_regression}
\end{figure}

\vspace{-10pt}
\subsubsection{Regression Task}
We evaluate the computational efficiency and predictive performance of \texttt{TorchCP} and an R-based CP package, \texttt{AdaptiveConformal}~\citep{susmann2023adaptiveconformal} on a time series regression task, using the France electricity forecasting dataset\footnote{\url{https://cran.r-project.org/web/packages/opera/index.html}}. Specifically, we use the first 350 time steps to train a SARIMA model with order \texttt{(1, 0, 0)} and seasonal order \texttt{(1, 1, 0)}, and construct prediction intervals for the remaining time steps. 
To assess computational efficiency, we apply the ACI algorithm\citep{gibbs2021adaptive} using both \torchcp and \texttt{AdaptiveConformal}.
Both \texttt{AdaptiveConformal} and \texttt{TorchCP} are executed on CPU in this experiment to ensure a fair comparison.
Moreover, since ACI performs recalibration for each test instance, the reported time includes both calibration and inference.

Figure~\ref{fig:consuming_time_regression} reports the runtime results of these two packages. The results show that \texttt{TorchCP} achieves significantly higher computational efficiency compared to the R-based package, reducing running time from 6.1 seconds to around 0.06 seconds. Figures~\ref{fig:AdConformal_prediction} and \ref{fig:torchcp_prediction} illustrate the prediction intervals produced by \texttt{AdaptiveConformal} and \torchcp, respectively. 
The two sets of intervals are nearly identical, demonstrating the correctness and consistency of the \torchcp implementation.  
This substantial improvement highlights the efficiency advantage of \torchcp for time series regression.

\section{The Advantage of our low-coupling design}
\label{appendix:dis_puncc}

Low-coupling design is one of the most fundamental concepts in object-oriented design, which minimizes the dependency a class has on other classes. The low-coupling design of \torchcp ensures flexibility in integrating diverse training algorithms, non-conformity score functions, and prediction workflows, accommodating the wide variety of CP methods and tasks. 
In contrast, most current packages focus on a high-coupling design where predictors and scoring functions are often predefined within a specific predictor (e.g., Split CP). This limits extensibility when new CP predictors or user-defined score functions are needed. \torchcp’s decoupled design allows users to independently select or define new trainers, scores, or predictors, enabling flexible integration of different components. In what follows, we elaborate on the concrete benefits of our low-coupling design:
\begin{itemize}
    \item \textbf{Flexible combination}: users can freely combine Trainer, Score, and Predictor modules to customize conformal prediction workflows. For instance, a practitioner can use different non-conformity scores (e.g., LAC or APS) within any predictor or trainer, while other packages only implement specified scores for predictors. Moreover, some libraries (e.g., \texttt{TorchUQ}) implemented various non-conformity scores as functions, but this style suffers from extending to those scores with extra hyperparameters (e.g., the $k$ and $\lambda$ of RAPS).
    \item \textbf{Modification isolation}: our low-coupling design ensures that changes to one module (e.g., refining the APS score) do not affect other modules. This isolation minimizes the risk of unintended side effects, reducing maintenance complexity and preventing errors from propagating across the system. For instance, tightly coupled systems like \texttt{PUNCC} require redefining the predictor to modify a scoring function, while \torchcp confines changes to the target module, streamlining updates and preserving stability.
    \item \textbf{Research extensibility}: by adhering to standardized interfaces, new functionality (e.g., a novel non-conformity score) can be integrated into \torchcp without altering existing modules. Differently, \texttt{MAPIE} follows a high-coupling design, making it challenging to introduce new predictors, like class-conditional CP. Extensibility supports rapid iteration and long-term evolvability, enabling efficient adaptation to new research or applications.
    \item \textbf{Ease of testing and debugging}: decoupled modules can be tested and debugged independently, as their interactions are governed by well-defined interfaces. This modularity simplifies identifying and resolving issues within a single module (e.g., debugging a Trainer algorithm) without navigating complex dependencies. In \torchcp, this reduces testing overhead and accelerates development cycles compared to tightly coupled systems, where debugging often requires analyzing the entire pipeline.
\end{itemize}

\section{Comparison with related libraries}
\label{appendix:comparsion_libraries}
Numerous libraries implement conformal prediction (CP) algorithms, yet their suitability for deep learning scenarios differs significantly, particularly in supporting modern deep learning models like GNNs and LLMs, as well as their scalability for large-scale datasets. In Table~\ref{tab:summarization}, we present a concise comparison of open-source CP libraries, focusing on supported model types and functionalities:
\begin{itemize}
    \item \textbf{Model type} refers to the types of machine learning models the library can work with. They are classified based on differences in input and output data, which influence the selection of applicable CP algorithms.
    \item \textbf{Functionality} describes the operational capabilities of the library, such as how it processes data, computational optimization, or prediction forms.
\end{itemize}
In the following, we compare several popular CP libraries, emphasizing TorchCP’s superior compatibility and performance with deep learning applications.

\textbf{nonconformist}~\citep{nonconformist}: A NumPy-based library supporting basic CP algorithms for classification and regression tasks. However, it lacks GPU acceleration, batch processing, and compatibility with deep learning models like DNN Classifiers, DNN Regressors, GNNs, or LLMs, making it unsuitable for complex deep learning scenarios. While \texttt{nonconformist} implements predictors and scores as different classes following a low-coupling design, its algorithmic coverage is limited, implementing only inductive CP and aggregated CP with 5 scores.

\textbf{TorchUQ}~\citep{torchuq2024}: A PyTorch library focused on uncertainty quantification, including some CP methods for regression, with GPU support. However, it lacks comprehensive algorithm implementation, omitting CP algorithms for classifiers and CP-specific training, and does not support advanced deep learning models like GNNs or LLMs, limiting its applicability for complex deep learning Scenarios. \texttt{TorchUQ} uses a dictionary to map names to score functions in predictors, enabling developers to define new scores without modifying predictor logic. However, this style struggles to accommodate scores with additional hyperparameters (e.g., $k$ and $\lambda$ in RAPS), requiring extra logic to manage these variations, which increases coupling and complexity.

\textbf{Crepes}~\citep{crepes}: a NumPy-based library with extensive prediction forms that supports the classifiers, regressors, and predictive systems with semi-online CP, excelling in statistical classification and regression tasks. However, it lacks GPU acceleration, batch processing, and seamless extensions to modern models like GNNs or LLMs, limiting scalability for engineers tackling large-scale or complex deep learning problems. Similar to \texttt{TorchUQ}, \texttt{Crepes} implements non-conformity scores as functions passed to predictors, so it also suffers from supporting scores with additional hyperparameters (e.g., $k$ and $\lambda$ in RAPS).

\textbf{PUNCC}~\citep{mendil2023puncc}: A scikit-learn-compatible library delivers Conformal Prediction for tasks like regression, classification, and anomaly detection, but struggles with modern deep learning applications. It does not support advanced deep learning models such as DNNs, GNNs, or LLMs, lacks GPU acceleration and batch processing, and performs significantly slower on large datasets (up to 10x slower than \torchcp in ImageNet, see Figure~\ref{fig:imagenet_consuming_time_diff}). \texttt{PUNCC} defines non-conformity scores as functions, but ties each score to a dedicated predictor, resulting in high coupling.

\textbf{MAPIE}~\citep{Cordier_Flexible_and_Systematic_2023}: A scikit-learn-based library providing CP sets/intervals for classical classification/regression tasks. It excels in traditional machine learning workflows but lacks support for deep learning models, such as DNNs, GNNs, or LLMs. It also does not offer online CP, GPU acceleration, and batch processing. \texttt{MAPIE} implements scores as different classes with a \texttt{predict} function, making it challenging to support other predictors (like class-conditional CP~\citep{vovk2012conditional}).

\textbf{Fortuna}~\citep{detommaso2024fortuna}: a JAX-based library, supports CP methods in deep learning with GPU acceleration and select conformal prediction methods. It enhances scalability for deep learning but has limited algorithmic coverage, lacking training algorithms and online CP, and does not support GNNs or LLMs. Additionally, JAX’s steeper learning curve may challenge statisticians familiar with PyTorch. Like \texttt{PUNCC}, \texttt{Fortuna} defines a dedicated classifier for each score, leading to high-coupling design.

In contrast, \torchcp, a PyTorch-native library, significantly enhances conformal prediction for deep learning by providing a comprehensive suite of state-of-the-art CP algorithms. It offers exceptional extensibility through low-coupling design with three modules (See Appendix~\ref{appendix:dis_puncc}) and unified interfaces across various algorithms (See Appendix~\ref{appendix:extensibility}). Notably, \torchcp seamlessly integrates CP algorithms with advanced deep learning models, such as GNNs and LLMs, and leverages GPU-accelerated batch processing to ensure scalability for large-scale deep learning scenarios (See Appendix~\ref{appendix:efficiency}).

\end{document}